\documentclass[sigconf,nonacm]{acmart}

\usepackage{amsmath}

\def\E{{\mathbb E}}
\def\R{{\mathbb R}}

\def\eps{{\varepsilon}}

\newtheorem{assumption}{Assumption}

\usepackage{algorithm}
\usepackage{algpseudocode}
\usepackage{enumitem}

\AtBeginDocument{%
  }

\begin{document}

\title{Optimization of Epsilon-Greedy Exploration}

\author{Ethan Che}
\affiliation{%
  \institution{Columbia Business School}
  \city{New York}
  \state{NY}
  \country{United States}
}
\author{Hakan Ceylan}
\affiliation{%
  \institution{Netflix}
  \city{Los Gatos}
  \state{CA}
  \country{United States}
  }
\author{James McInerney}
\affiliation{%
  \institution{Netflix}
  \city{Los Gatos}
  \state{CA}
  \country{United States}
}
\author{Nathan Kallus}
\affiliation{%
 \institution{Netflix \& Cornell University}
 \city{New York}
  \state{NY}
  \country{United States}
}

\renewcommand{\shortauthors}{Che et al.}

\begin{abstract}
Modern recommendation systems rely on exploration to learn user preferences for new items, typically implementing uniform exploration policies (e.g., epsilon-greedy) due to their simplicity and compatibility with machine learning (ML) personalization models. Within these systems, a crucial consideration is the \emph{rate of exploration} - what fraction of user traffic should receive random item recommendations and how this should evolve over time. While various heuristics exist for navigating the resulting exploration-exploitation tradeoff, selecting optimal exploration rates is complicated by practical constraints including batched updates, time-varying user traffic, short time horizons, and minimum exploration requirements. In this work, we propose a principled framework for determining the exploration schedule based on directly minimizing Bayesian regret through stochastic gradient descent (SGD), allowing for dynamic exploration rate adjustment via Model-Predictive Control (MPC). Through extensive experiments with recommendation datasets, we demonstrate that variations in the batch size across periods significantly influence the optimal exploration strategy. Our optimization methods automatically calibrate exploration to the specific problem setting, consistently matching or outperforming the best heuristic for each setting.
\end{abstract}





\maketitle

\section{Introduction}


Modern recommendation systems apply sophisticated machine learning (ML) algorithms to historical interaction data to determine optimal personalized recommendations for each user. This shift has substantially improved recommendation quality but introduces significant challenges when new items or treatments enter the system.
The introduction of new items creates the well-documented "cold-start" problem in recommendation systems~\cite{schein2002methods}, where no historical data exists to inform personalization decisions. While numerous techniques have been proposed for extrapolating from existing items to new ones using content features or collaborative signals, a prevalent practice remains collecting new interaction data through dedicated exploration phases~\cite{li2010contextual, nguyen2014cold, chen2021exploration}. This approach creates a fundamental tension: more exploration yields better training data for recommendation algorithms but delays the delivery of optimally personalized experiences to users. This tension represents the classic "exploration-exploitation" trade-off extensively studied in both recommendation systems and the broader bandit literature~\cite{lattimore2020bandit, barraza2017exploration}.

Although sophisticated algorithms from the contextual bandit literature such as Thompson Sampling (TS)~\cite{russo2018tutorial} are theoretically shown to be effective at managing this trade-off, production recommendation systems often implement simpler exploration strategies based on \emph{uniform exploration} of items due to (1) engineering and operational constraints~\cite{mcinerney2018explore} and (2) compatibility with black-box machine learning models for personalization. Two particularly common approaches in recommendation platforms are:
\begin{itemize}[leftmargin=0.5cm]
\item \textbf{Explore-then-commit (ETC):}~\cite{garivier2016explore} This strategy dedicates an initial period to randomly assigning treatments across the user base before switching to a fully personalized policy.
\item \textbf{Epsilon-greedy:}~\cite{sutton2018reinforcement} This approach continuously allocates a fixed percentage of traffic to random items while using the current ML policy for the remainder.
\end{itemize}
A critical challenge for recommendation system practitioners implementing these strategies is determining the optimal rate of exploration -- how many users will be allocated a random item instead of the item recommended by the ML algorithm? For ETC, this means selecting the appropriate duration of the initial exploration phase. For epsilon-greedy, it requires setting the right percentage of traffic devoted to exploration. While theoretical guidelines exist in the bandit literature~\cite{vermorel2005multi, lattimore2020bandit}, real-world recommendation systems face additional complexities that theoretical models often overlook:

\begin{itemize}[leftmargin=0.5cm]
\item \textbf{Batched updates:}~\cite{van2024practical} Unlike the fully sequential sampling assumed in theoretical work, real recommendation systems typically collect data in batches and update models periodically (e.g., daily), introducing latency between exploration and policy improvement~\cite{perchet2016batched, gao2019batched, esfandiari2021regret}.
\item \textbf{Time-varying user traffic:} Recommendation platforms experience fluctuating traffic patterns, especially when new items are released (e.g. new movies). Thus, the batch sizes can vary substantially across time periods, affecting the rate at which exploration data accumulates.
\item \textbf{Limited horizons:} New content launches or seasonal recommendations often operate within constrained timeframes (1-2 weeks), requiring rapid convergence to optimal policies before user interest wanes.
\item \textbf{Exploration constraints:} Minimum exploration thresholds may be necessary to enable reliable statistical inference or counterfactual evaluation of recommendation policies.
\end{itemize}

These operational realities of recommendation systems significantly complicate the exploration strategy. As a result, exploration parameters are frequently determined through ad-hoc approaches or historical precedent rather than principled optimization. Despite the maturity of the recommendation systems field, there remains a notable gap between theoretical exploration strategies and practical implementation guidance for these constrained, real-world recommendation scenarios.


In this work, we introduce a principled framework for optimizing exploration in limited horizon, batched settings. Our approach leverages a Bayesian model and formulates exploration as an explicit optimization problem, in which the exploration rates are chosen to minimize Bayesian regret. In summary our contributions are as follows:
\begin{itemize}[leftmargin=0.5cm]
\item Using a Bayesian model, we obtain a \emph{differentiable} formulation of Bayesian regret as a function of the exploration rates that can be directly minimized through stochastic gradient descent.
\item By minimizing regret via SGD, we solve for a schedule of exploration rates tailored to the problem setting. By re-solving the optimization problem after every batch--a control policy known as Model-Predictive Control (MPC)--the platform can adjust the exploration rate dynamically.
\item Through extensive computational experiments, we find that batched updates significantly affect the optimal rate of exploration, and depending on the structure of the batches, simple rules-of-thumb can outperform theoretically-derived guidelines for the exploration schedule.
\item We find that our proposed methods are able to adapt the exploration schedule precisely to the structure of the batched arrivals, and match or outperform the best heuristic for each setting.
\item We find surprisingly that in these short-horizon, batched settings, uniform exploration with an appropriate exploration schedule can even outperform more sophisticated approaches such as Batched Thompson Sampling~\cite{kalkanli2021batched, karbasi2021parallelizing}.
\end{itemize}
While we focus on the linear-contextual bandit setting, our results can be extended to more complex models.

The paper is organized as follows. We discuss related work on exploration and contextual bandits in recommendation systems in Section~\ref{sec:related_work}. In Section~\ref{sec:setting}, we introduce the setting and the contextual bandit problem. In Section~\ref{sec:mpc}, we describe the Bayesian formulation and the regret objective. In Section~\ref{sec:opt}, we discuss the optimization problem and the model-predictive control algorithm. In Section~\ref{sec:experiments}, we benchmark our proposed algorithms with baseline exploration strategies on several datasets. Finally, conclusions and future work are discussed in Section~\ref{sec:conclusion}.

\section{Related Work}
\label{sec:related_work}

Our work is closely related to bandits for personalization and recommendation as well as Bayesian optimal experimental design. 
\paragraph{Bandits for Recommendation Systems} Multi-armed and contextual bandits are a popular methodology for cold-start exploration and personalization~\cite{tang2014ensemble,chen2021exploration, barraza2017exploration, mcinerney2018explore, mary2014bandits, silva2023user, glowacka2019bandit}. Rather than designing a new bandit policy, we focus on existing uniform exploration policies, due to their widespread usage~\cite{mcinerney2018explore}, and develop a framework to tune exploration rates to handle practical constraints such as batched/infrequent feedback~\cite{van2024practical}.

\paragraph{Bayesian Optimal Experimental Design}
Our methods rely on a Bayesian model to calibrate exploration for personalization, similar to~\cite{wang2014exploration}. Our approach, which formulates exploration as an optimization problem, is closely related to Bayesian methods for optimal experimental design~\cite{rainforth2024modern, foster2021deep} and Bayesian methods for regret minimization~\cite{che2023adaptive,che2024optimization}. However, while typically these works typically focus on information gain or minimizing posterior uncertainty, our objective is Bayesian regret and we devise efficient approximations for optimizing this. Furthermore, our scope is limited to tuning the rate of exploration, rather than experimental selection.

\begin{figure}
\includegraphics[width=0.7\columnwidth]{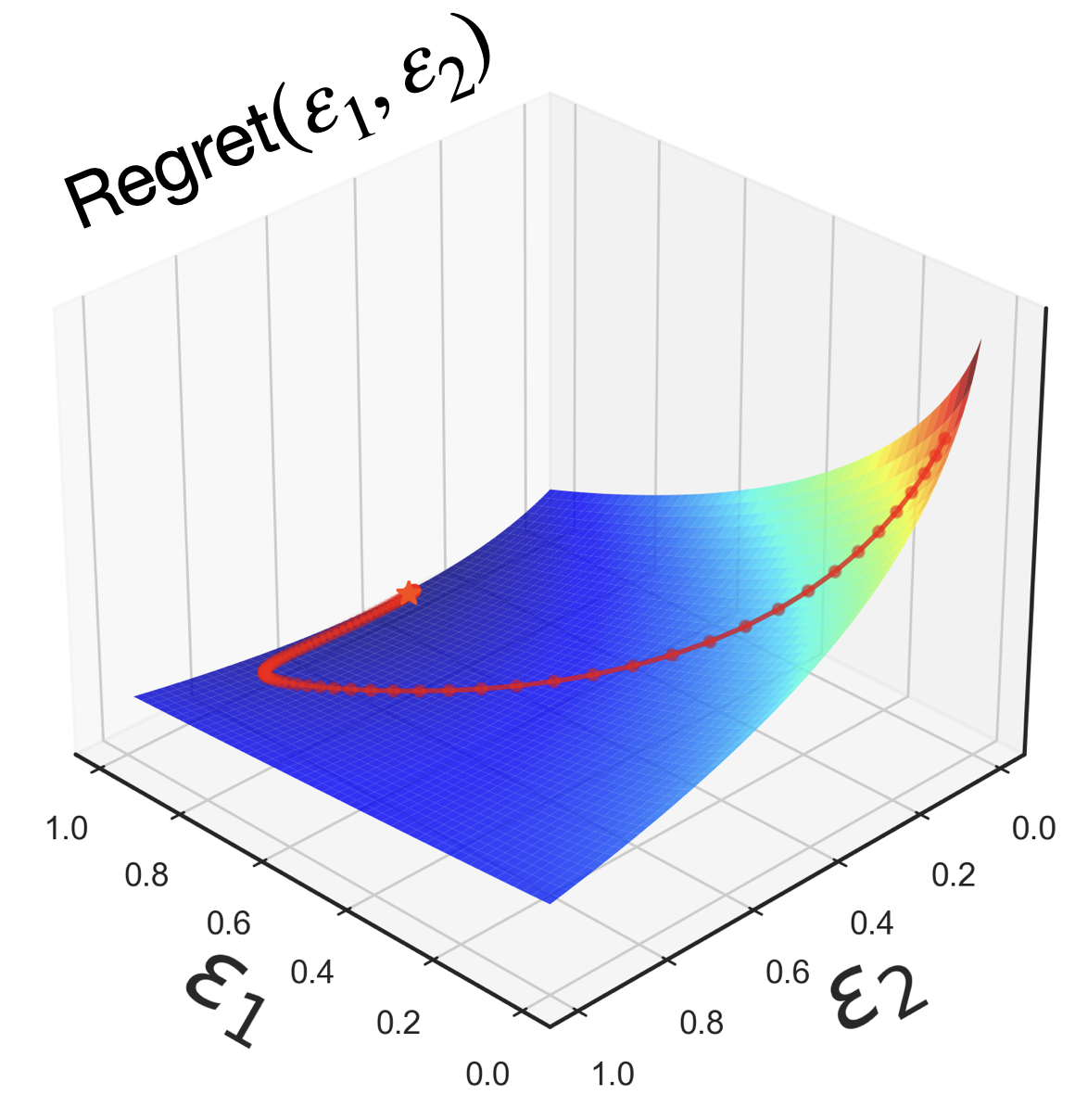}
\caption{Our proposed method selects the exploration rates for each period $\eps_{t} \in [0,1]$  by choosing the rates that minimize a differentiable approximation of Bayesian Regret (displayed as a function of the exploration rates in periods 1 and 2, $\eps_{1}$ and $\eps_{2}$). Due to differentiability, regret can be efficiently minimized with stochastic gradient descent (in red, see Algorithm~\ref{alg:gd}).}
\end{figure}
\section{Setting}
\label{sec:setting}

We consider the problem of selecting the exploration rate when collecting data to train a personalized assignment policy on newly-released items over a fixed time horison. Following convention, we model this as a batched contextual bandit problem, which we proceed to formalize.

We denote $\mathcal{A}$ to be a set of $K \in \mathbb{N}$ newly-released items. There are $T\in \mathbb{N}$ time periods and in each period $t = 1,...,T$ a batch of $n_{t} \in \mathbb{N}$ users arrive to the platform, with a total of $n = \sum_{t=1}^{T} n_{t}$ users. For example, each period $t$ could correspond to a day, in which case $n_{t}$ is number of impressions that arrive on day $t$. Each user has a user embedding $X_{i}\in \R^{d}$, and we assume that in each batch users arrive iid from a distribution $X_{i} \sim \mu $. We assume that the user embeddings are $X_{i}$ are frozen and are not being learned simultaneously. 

In each period $t$, after the batch of $n_{t}$ users arrives, the platform assigns each user $i$ a \emph{single} item $A_{i} \in \mathcal{A}$ according to a policy $\pi_{t}$, which is a probability distribution over $\mathcal{A}$ conditional on the user embedding $X_{i}$, i.e. $A_{i} \sim \pi_{t}(\cdot |X_{i})$.
After assigning an item to the user, the platform  observes a reward metric $R_{i}$, e.g. engagement or user rating. We consider a standard form of the reward model, which posits reward as a linear function of the user embedding:
\begin{align}
\label{eqn:reward}
r(x,a) \equiv  x^\top \theta _{a}
\end{align}
where $\theta_{a}\in \mathbb{R}^{d}$ is an item embedding that is unknown to the platform. As a result, this problem is a linear contextual bandit problem. We refer to $\theta \in \R^{Kd}$ as the vector containing the embeddings for all $K$ items. The platform only observes a noisy reward:
\begin{align}
\label{eqn:stoch_reward}
R_{i} \equiv R_{i}(A_{i}) = r(X_{i}, A_{i}) + \eta_{i}
\end{align}
where $\eta_{i}$ are iid random variables with zero mean $\mathbb{E}[\eta_{i}] = 0$ and variance $\text{Var}(\eta_{i})=s^{2}$. After collecting data $D_{t} = \{(X_{i}, A_{i},R_{i})\}_{i=1}^{n_{t}}$ from batch $t$, the platform updates their estimate for $\theta$, denoted as $\hat{\theta}_{t}$, and their policy $\pi_{t+1}$ for the next batch.

Crucially, due to the batched nature of the feedback, the platform only updates their policy after observing all the rewards at the end of the batch. This is designed to match practical settings for which the personalization algorithm is updated after observing a large batch of data (e.g. all the interactions during a day), instead of being updated after each sample, as is the standard assumption in the contextual bandit literature. 

A commonly considered goal is to find a policy that minimizes expected cumulative regret along the time horizon:
\begin{align}
\label{eqn:regret}
    \min_{\pi}  \text{Regret}(\pi) \equiv \min_{\pi}  \sum_{t=1}^{T} \sum_{i=1}^{n_{t}} \E_{X_{i}\sim \mu}\left[ \text{Reg}(X_{i}, \pi_{t}) \right]
\end{align}
where the regret for user $i$ is defined to be the difference between the expected reward under the best item and the expected reward achieved under the policy,
\begin{align}
    \text{Reg}(X_{i},\pi_{t}) &\equiv r^{*}(X_{i}) - \E_{A_{i} \sim \pi_{t}(\cdot |X_{i})} [r(X_{i},A_{i})] \\
    r^{*}(X_{i}) &\equiv \max_{a\in\mathcal{A}} \{ r(X_{i},a) \}
\end{align}

While there exist policies, such as (batched) Thompson Sampling, that are designed to achieve sublinear regret~\eqref{eqn:regret} as the number of users $n$ grows large, we focus on the class of \emph{uniform exploration policies}, motivated by the widespread use in production recommendation systems. Due to the simplicity of their design, uniform exploration policies are easy to implement, work well with batched feedback and other production constraints, and are compatible with more complex reward models.

At each period $t$, the policy randomly allocates a fraction of users to a uniformly sampled item. We denote this fraction as the \emph{exploration rate} $\eps_{t}$ and refer to this group as the `explore' group. We let $\xi_{i}\in\{0,1\}$ indicate whether $i$ is in the explore group. The remaining $1-\eps_{t}$ are assigned an item `greedily' based on the current estimate $\hat{\theta}_{t}$ of $\theta$ (the `exploit' group). As a result, the policy will be of the following form:
\begin{align}
\label{eqn:uniform_policy}
\pi_{t}(a|X_{i}) =
\begin{cases}
1/K & \text{w.p. }  \eps_{t} \\
\mathbf{1}
\{
a = \arg \max_{a} X_{i}^{\top}\hat{\theta}_{t,a}
\}
& \text{w.p. } 1- \eps_{t}
\end{cases}
\end{align}
Restricting ourselves to policies of this form, a key question is determining the proper exploration rates $\eps_{1:T} \equiv \{\eps_{t}\}_{t=1}^{T}$. We can express regret as a function of the exploration rates (abusing notation):
\begin{align}
\label{eqn:eps_regret}
    \text{Regret}(\eps_{1:T})& \equiv   \sum_{t=1}^{T} \sum_{i=1}^{n_{t}} \E_{X_{i}\sim \mu}\left[ \text{Reg}(X_{i}, \eps_{t}) \right] \\
    \text{Reg}(X_{i},\eps_{t}) &\equiv 
     \eps_{t}\text{Reg}(X_{i}, \mathbf{1}/K)  
     + (1-\eps_{t})\text{Reg}(X_{i}, \widehat{\pi}_{t})\\
    \widehat{\pi}_{t}(a|X_{i}) &\equiv 
    \mathbf{1}
\{
a = \arg \max_{a} X_{i}^{\top}\hat{\theta}_{t,a}
\}
\end{align}
where $\text{Reg}(X_{i}, \mathbf{1}/K)$ is the regret under a uniformly drawn item and $\text{Reg}(X_{i}, \widehat{\pi}_{t})$ is regret under the greedy assignment of the item based on the current estimate $\hat{\theta}_{t}$. 

Two well-known examples of uniform exploration policies are:
\begin{itemize}
    \item Epsilon-greedy uses a constant exploration rate $\eps_{t} = \eps$.
    \item Explore-then-Commit (ETC) sets a fixed budget of $B$ samples before committing to the greedy policy. Thus, $\eps_{t} = 1$ until the budget is reached after which $\eps_{t} = 0$:
    \begin{align}
        \eps_{t} = 
         \min\left\{ \frac{(B - \sum_{s=1}^{t-1}\eps_{s}n_{s})} {n_{t}}, 1 \right\} 
    \end{align}
    Note that $\eps_{t}$ may be fractional if $\sum_{s=1}^{t-1}n_{s} <  B < \sum_{s=1}^{t}n_{s}$.
\end{itemize}

While generally a larger exploration rate $\varepsilon_{t}$ item will incur more regret in period $t$, as typically assigning a random item has higher regret $\text{Reg}(X_{i}, \mathbf{1}/K) > \text{Reg}(X_{i}, \widehat{\pi}_{t})$, exploration through random item assignment improves the estimate $\hat{\theta}_{t}$ for future periods -- this presents the classic `exploration-exploitation' trade-off.
However, it is not clear how to set the exploration rates in order to precisely optimize this trade-off, especially in batched settings with (potentially) time-varying batch sizes $n_{t}$. And while there exist theoretically-derived exploration rates (e.g. in~\cite{garivier2016explore, lattimore2020bandit}), they often involve either unknown quantities (e.g. gaps between average rewards) or under-specified constants that can impact practical performance. As a result, existing practices for selecting $\eps_{t}$ are typically ad-hoc.

\section{Differentiable Bayesian Regret}
\label{sec:mpc}

Ideally, we could select the exploration rates $\eps_{1:T}$ to directly minimize $\text{Regret}_{T}(\eps_{1:T})$ in~\eqref{eqn:eps_regret}. However, there are two issues that make $\text{Regret}_{T}(\eps_{1:T})$ unsuitable for optimization. First, $\text{Regret}_{T}(\eps_{1:T})$ depends on the unknown item embeddings $\theta$ through the reward 
model $r(x,a)$. Second, quantifying the benefit of exploration requires precisely understanding how additional samples will affect the greedy policy in future periods, which is difficult to characterize and does not have a simple closed form that can be optimized. In general, it is well-known that regret  is a difficult objective to directly optimize, which is why existing algorithmic principles such as UCB or Thompson Sampling rely on optimizing proxies or upper bounds of regret instead (e.g. Hoeffding Bound)~\cite{lattimore2020bandit}. However, in this section, we show that \emph{Bayesian regret} does not suffer from these two issues and is amenable to optimization.

We assume the platform has a prior distribution over the unknown item embeddings:
\begin{align}
\theta \sim N(\beta_{1}, \Sigma_{1})
\end{align}
where $\beta_{1}\in \R^{Kd}$ and $\Sigma_{1} \in \R^{Kd \times Kd}$ are the prior mean and covariance of the item embeddings. Typically, $\beta_{1} = \mathbf{0}$ and $\Sigma = \sigma^{2}\mathbf{1}$ for some $\sigma^{2}>0$, although the prior could be calibrated to the statistics of existing item embeddings. 

We introduce three assumptions that enable simplify the structure of the Bayesian posterior updates. First, We assume Gaussian noise, as this implies that the posterior distribution of $\theta$ is also Gaussian with closed-form expressions for the posterior mean and covariance. 
\begin{assumption}
\label{ass:normal}
The noise $\eta_{i}$ are distributed iid and are normally distributed $N(0,s^{2})$.
\end{assumption}
We also assume that posterior updates only use data from the `explore' group:
\begin{assumption}
\label{ass:explore_only}
The posterior is updated using only data from the `explore' group, i.e. the history we are conditioning on $H_{t} = \sigma(\cup_{s=1}^{t-1}\mathcal{D}_{s})$ where $\mathcal{D}_{s} = \cup_{i=1}^{n_{s}}\{(X_{i},A_{i},R_{i},\xi_{i}):\xi_{i}=1\}$.
\end{assumption}
One limitation of this assumption is that it ignores data gathered by the greedy policy. Nevertheless this captures the exploration-exploitation tradeoff and reflects practical implementations, which typically only use randomly sampled data to train the policy.

Our final assumption is that the prior covariance $\Sigma_{1}$ is block-diagonal across items. 
\begin{assumption}
\label{ass:block}
We assume that the prior covariance is of the form:
    $$\Sigma_{1} = \mathbf{diag}(\Sigma_{1,1},\ldots \Sigma_{1,a}, \ldots, \Sigma_{K,a})$$
where $\Sigma_{1,a} \in \R^{d\times d}$ for all $a\in \mathcal{A}$.
\end{assumption}
We do not need this assumption for the validity of Bayesian updating, but it simplifies notation by allowing for the posterior to be updated independently for each item.

Under these assumptions, we obtain a standard expression for the posterior distribution of $\theta$ at the beginning of period $t$. The posterior distribution of $\theta$ given $H_{t}$ follows a normal distribution:
\[
\theta|H_{t} \sim N(\beta_{t},\Sigma_{t})
\]
where the posterior mean $\beta_{t}$ and covariance $\Sigma_{t}$ follow the recursive formula for each $a$:
\begin{align}
    \Sigma_{t+1,a} &= \left( \Sigma_{t,a}^{-1} + I_{t,a}(\eps_{t}) \right)^{-1} \label{eqn:true_sigma_update}\\
    \beta_{t+1,a} &= \Sigma_{t+1,a}\left( \Sigma_{t,a}^{-1} \beta_{t} + I_{t,a}(\eps_{t})^{-1} \left(\sum_{i=1}^{n_{t}}  \xi_{i}\mathbf{1}\{A_{i} = a\}R_{i}X_{i} \right) \right) \label{eqn:true_beta_update}\\
    I_{t,a}(\eps_{t})  & \equiv s^{-2}\sum_{i=1}^{n_{t}}\xi_{i}\mathbf{1}\{A_{i} = a\}X_{i}X_{i}^{\top} \label{eqn:stoch_design}
\end{align}

We use the notation $\E_{1}$ to denote expectation of $\theta$ under the prior and $\E_{t}[\cdot]\equiv \E[\cdot |H_{t}]$ to denote expectation under the posterior at period $t$ so,
\begin{align}
    \E_{t}[r(x,a)]= x^\top \E_{t}[\theta_{a}]=x^\top \beta_{t,a}
\end{align}


An important property of the Bayesian linear regression model for optimization is the existence of closed-form expressions for the distribution of \emph{future} posterior states, conditional on the current posterior and a fixed exploration rate. A derivation can be found in~\cite{che2024optimization}.

\begin{lemma}~\cite{che2024optimization}
\label{lemma:posterior}
Suppose Assumptions~\ref{ass:normal}-\ref{ass:block} hold. Consider any exploration rates $\eps_{1:T}$, and any $1\leq t\leq s$. For each item $a$, the distribution of the posterior $(\beta_{s,a}, \Sigma_{s,a})$, conditional on the period $t$ posterior $(\beta_{t,a}, \Sigma_{t,a})$ is given as follows:
\begin{align}
    \Sigma_{s,a} &\stackrel{d}{=} \left( \Sigma_{t,a}^{-1} +  \sum\nolimits_{l=t}^{s-1}I_{t,a}(\eps_{t}) \right)^{-1} \label{eqn:pred1}\\
    \beta_{s,a} &\stackrel{d}{=} \beta_{t,a} + (\Sigma_{t,a} - \Sigma_{s,a})^{1/2}Z_{s} \label{eqn:pred2}
\end{align}
where $Z_{t} \sim N(0,I)$ is a standard Gaussian random vector and $I^{a}_{n,s}$ is the empirical design matrix given by~\eqref{eqn:stoch_design} with $X_{i}\stackrel{iid}{\sim}\mu$ ($\mu$ is the distribution of user embeddings), $\xi_{i}\stackrel{iid}{\sim}\text{Ber}(\eps_{s})$, $A_{i}\stackrel{iid}{\sim}\text{Multinomial}(\mathbf{1}/K)$.
\end{lemma}
Lemma~\ref{lemma:posterior} captures how future posterior beliefs will change based on the exploration rates. Setting a high exploration rate $\eps_{t}$ induces a larger update in future beliefs while a low exploration rate will not change future beliefs from the current posterior. Thus, the lemma precisely measures the \emph{value of exploration} not only for the next period but for every future period in the time horizon, and doing so only requires sampling from a standard normal distribution.

We now turn to the Bayesian regret objective, which is more amenable to optimization as it does not depend on unknown item embeddings. Instead, it averages regret~\eqref{eqn:eps_regret} over item embeddings drawn from the prior distribution:
\begin{align}
& \min_{\eps_{1:T}}   \E_{\theta \sim N(\beta_{1},\Sigma_{1} )}\left[\text{Regret}_{T}(\eps_{1:T}) \right] \label{eqn:bayes_eps_regret}\\
&=\min_{\eps_{1:T}}
\sum_{t=1}^{T} n_{t} \E_{1}[
\eps_{t}\text{Reg}(X_{i}, \mathbf{1}/K)  
     + (1-\eps_{t})\text{Reg}(X_{i}, \widehat{\pi}_{t})]
\end{align}
where the expectation in the second line denotes expectation over $X_{i}\sim \mu$, $\xi_{i} \sim \text{Ber}(\eps_{i})$, random sampling $A_{i}$, and the prior distribution of $\theta$. One benefit of the Bayesian model is that Regret can be written in closed form in terms of prior and posterior parameters. To illustrate, we first look at the regret of the greedy policy $\text{Reg}(X_{i}, \widehat{\pi}_{t})$. Given the Bayesian formulation, we slightly revise the definition of the greedy policy $\hat{\pi}_{t}$: it assigns the item that maximizes expected reward at period $t$,
\begin{align}
    \hat{\pi}_{t}(a|x)& \equiv
    \mathbf{1}\{ a = \arg\max_{a} \E_{t}[r(X_{i},a)]\} \label{eqn:bayes_greedy} \\
    &=\mathbf{1}\{ a = \arg\max_{a} X_{i}^{\top}\beta_{t,a}\}.  \nonumber
\end{align}

We arrive at our main result, which gives a concrete expression for Bayesian regret.
\begin{proposition}
\label{thm:diff_regret}
Under Assumptions~\ref{ass:normal}-\ref{ass:block}, Bayesian regret~\eqref{eqn:bayes_eps_regret} under the uniform exploration policy with the greedy assignment given by~\eqref{eqn:bayes_greedy} is equal to 
\begin{align}
J_{1}(\eps_{1:T},\beta_{1},\Sigma_{1}) &\equiv \E_{1}[\text{Regret}_{T}(\eps_{1:T})] \label{eqn:diff_regret}  \\
&= n\E_{1}[r^{*}(X_{i})]-
\sum_{t=1}^{T}
n_{t}
\E\left[\eps_{t} X_{i}^{\top}\bar{\beta}_{1} + (1-\eps_{t})\max_{a}X_{i}^{\top}\beta_{t,a}\right] \nonumber
\end{align}
where $\bar{\beta}_{1} = (1/K)\sum_{k=1}^{K} \beta_{1,a}$ and for all items $a\in \mathcal{A}$,
\begin{align}
\Sigma_{t,a}^{-1} &= \Sigma_{1,a}^{-1} +  \sum\nolimits_{l=1}^{t-1}I_{l,a}(\eps_{l})
\label{eqn:sigma_update}
\\
    \beta_{t,a} &= \beta_{1,a} + (\Sigma_{1,a} - \Sigma_{t,a})^{1/2}Z_{t} 
    \label{eqn:beta_update}
\end{align}
where $I_{t,a}(\eps_{t})$ is defined in~\eqref{eqn:stoch_design} and the expectation in~\eqref{eqn:diff_regret} is over $Z_{t}\stackrel{iid}{\sim}N(0,I)$, $X_{i}\stackrel{iid}{\sim}\mu$, $\xi_{i}\stackrel{iid}{\sim}\text{Ber}(\eps_{s})$, and $A_{i}\stackrel{iid}{\sim}\text{Multinomial}(\mathbf{1}/K)$.
\end{proposition}

See proof in Section~\ref{sec:appendix}. Proposition~\ref{thm:diff_regret} provides an exact expression for Bayesian regret, which can be efficiently estimated through Monte Carlo simulation of standard random variables based on known information. It presents a precise characterization of the exploration-exploitation trade-off -- a higher $\eps_{t}$ reduces immediate reward by placing more weight on randomization, but increases future reward by increasing the variance of future posteriors $\beta_{t,a}$ (i.e. inducing a larger posterior update), which increases the expected reward from the greedy policy in the future $\E_{1}[\max_{a}X_{i}^\top \beta_{t,a}]$. 

Still, the expressions~\eqref{eqn:sigma_update}-\eqref{eqn:beta_update} are difficult to directly optimize because $I_{t,a}(\eps_{t})$ depends on $\eps_{t}$ through the realizations of Bernoulii r.v.s. $\xi_{i}$. Nevertheless, when $n_{t}$ is large the dependence of $I_{t,a}$ on $\eps_{t}$ simplifies, as a consequence of the Law of Large Numbers. To formalize this, we introduce a mild assumption bounding the norm of the user embeddings.
\begin{assumption}
\label{ass:bounded}
$\|X_{i}\|^{2}_{2} \leq C$ almost surely.
\end{assumption}
Given that there are a finite number of users, this is generally satisfied. We proceed to show that for large batch sizes, the empirical design matrix converges to a population design matrix, using a standard matrix concentration inequality.
\begin{lemma}
\label{lemma:empirical}
    Suppose Assumption~\ref{ass:bounded} holds and $\eps_{t}>0$ for all $t$. Moreover, suppose that $n_{t}=\lambda_{t}n$ for some $\lambda_{t} \in (0,1)$ where $\mathbf{1}^\top\lambda = 1$. Then for any $t\leq s\leq T$, for all $\delta >0$, with  probability at least $1-\delta$ we have that as $n\to \infty$
    \[
    \left\Vert\sum\nolimits_{l=t}^{s} I_{t,a}(\epsilon_{t}) - \sum\nolimits_{l=t}^{s}n_{l}\eps_{t}I \right\Vert_{\text{op}} =O(n^{1/2}\log(1/\delta))
    \]
    where $I$ is the population design matrix.
\begin{align}
\label{eqn:design}
\mathcal{I}\equiv s^{-2}K^{-1}\E_{X_{i}\sim\mu}[X_{i}X_{i}^{\top}]
\end{align}
\end{lemma}
\noindent See Appendix~\ref{sec:appendix} for the proof.


\begin{algorithm}[t]
\caption{Solve $\min_{\eps_{t:T}} \overline{J}_{t}(\eps_{t:T}, \beta_{t},\Sigma_{t})
$ via SGD}\label{alg:gd}
\begin{algorithmic}
\Require Current period $t$, current posterior $(\beta_{t}, \Sigma_{t})$, user embedding samples $\{X_{i} \}_{i=1}^{m}$, design matrix $\mathcal{I}$, batch sizes $n_{t}$, number of gradient steps $L$, step size $\alpha$, feasible domain $\Omega \subseteq [0,1]^{T-t-1}$.
\Ensure Exploration rates $\eps_{t:T}$

\State Initialize $\eps_{t:T}^{(0)} \in [0,1]^{T-t-1}$.
\For{$k = 0, \ldots L-1$} \Comment{$L$ steps of gradient descent}
\State Draw iid $\{Z_{s}\}_{s=t}^{T}$ where $Z_{s}\sim N(0,I).$
\For{$s = t,\ldots,T$} \Comment{Simulate posterior from $t$ to $T$}
\State $\Sigma_{s}^{-1} =\Sigma_{t,a}^{-1} +  \sum\nolimits_{l=1}^{s-1}\eps^{(k)}_{l}n_{l}\mathcal{I}$
\Comment{Covariance update}
\State $\beta_{s,a} = \beta_{t,a} + (\Sigma_{t,a} - \Sigma_{s,a})^{1/2}Z_{s}$ \Comment{Mean update}

\EndFor
\State $J(\eps_{t:T}) \gets -\sum_{s=t}^{T} \frac{1}{m}\sum_{i=1}^{m} 
\left( \eps_{s} X_{i}^{\top}\bar{\beta}_{1} + (1-\eps_{s})\max_{a}X_{i}^{\top}\beta_{s,a} \right)$
\State Compute gradient $\nabla J(\eps_{t:T})$ wrt $\eps_{t:T}$.
\State $\eps_{t:T}^{(k+1)} \gets \text{Proj}_{\Omega}(\eps_{t:T}^{(k)} - \alpha \nabla J(\eps_{t:T}))$

\EndFor
\end{algorithmic}
\end{algorithm}

The population design matrix is more suitable for optimization, as the design matrix no longer depends on $\eps_{t}$ through the realization of Bernoulli r.v.s and instead is simply linear in $\eps_{t}$. We substitute into~\eqref{eqn:sigma_update} to obtain the following approximation for the update for $\Sigma$, and we denote the resulting covariance matrices as $\overline{\Sigma}_{t,a}$
\begin{align}
\label{eqn:diff_sigma_update}
\overline{\Sigma}_{t,a}^{-1} &= \Sigma_{1,a}^{-1} +  \sum\nolimits_{l=1}^{t-1}\eps_{l}n_{l}\mathcal{I}
\end{align}
Let $\overline{J}_{1}(\varepsilon_{1:T},\beta_{1},\Sigma_{1})$ be identical to $J_{1}(\varepsilon_{1:T},\beta_{1},\Sigma_{1})$ in~\eqref{eqn:diff_regret}, but replacing $\Sigma_{t,a}$ with $\overline{\Sigma}_{t,a}$ for all $t,a$. Since $\overline{\Sigma}_{t,a}$ is differentiable in $\eps_{1:t-1}$, $\overline{J}_{1}$ serves as a fully differentiable approximation of Bayesian cumulative regret $J$. This approximation is also easier to simulate and only requires samples of $Z_{t}\stackrel{iid}{\sim}N(0,I)$, $X_{i}\stackrel{iid}{\sim}\mu$.

We turn to our second main result, which shows that this approximation is accurate as $n\to \infty$
\begin{theorem}
\label{thm:approximation}
Suppose Assumption~\ref{ass:bounded} holds, $n_{t}=\lambda_{t}n$ as in Lemma~\ref{lemma:empirical}, and $\eps_{t}>0$ for all $t$. For all $\delta >0$, with probability at least $1-\delta$ we have as $n\to \infty$,
\begin{align}
|J_{1}(\varepsilon_{1:T},\beta_{1},\Sigma_{1}) - \overline{J}_{1}(\varepsilon_{1:T},\beta_{1},\Sigma_{1})|=O(n^{1/4}\log(1/\delta)).
\end{align}
\end{theorem}
\noindent See Appendix~\ref{sec:appendix} for the proof.

The above bound for Bayesian cumulative regret implies that the discrepancy in \emph{average} Bayesian regret decays at rate $O(n^{-3/4})$ with high probability. Not only is the approximation error sublinear in $n$, since cumulative regret is known to scale at least $\Omega(n^{1/2})$~\cite{chu2011contextual}, this result shows that the approximation error is a much smaller order than regret. Thus, for large sample sizes, $\overline{J}$ can serve as a useful approximation for Bayesian cumulative regret, which is suitable for optimization via gradient-based methods. This result also extends to $J_{t}(\eps_{t:T},\beta_{t},\Sigma_{T})$ and $\overline{J}_{t}$, which refer to the cumulative regret starting from period $t$ and posterior $(\beta_{t}, \Sigma_{t})$. 

\section{Exploration via Optimization}
\label{sec:opt}





Bayesian regret offers a suitable optimization objective as (1) it elicits a differentiable approximation that is accurate under large batches and (2) Monte-Carlo simulation of this approximation can be efficiently generated by simulating $Z_{i}\sim N(0,I)$ and drawing a sample of user embeddings $X_{i}\sim \mu$.

We proceed to introduce the optimization problem for determining the exploration rates. Note that the optimal reward $n\E_{1}[r^{*}(X_{i})]$ is not affected by $\eps$, and so minimizing regret is equivalent to maximizing the reward obtained by the policy. This leads to the following stochastic optimization problem: let $Z_{t}\sim N(0,I)$ be iid standard normal vectors,
\begin{align}
\label{eqn:opt}
\begin{array}{rl}
\max\limits_{\eps_{1:T}} & \sum_{t=1}^{T}
n_{t}
\E\left[\eps_{t} X_{i}^{\top}\bar{\beta}_{1} + (1-\eps_{t})\max_{a}X_{i}^{\top}\beta_{t,a}\right] \\
\text{s.t.} & \Sigma_{t,a}^{-1} = \Sigma_{1,a}^{-1} +  \sum\nolimits_{l=1}^{t-1}\eps_{l}n_{l}\mathcal{I}, \quad \forall t=1,...,T \\
            & \beta_{t,a} = \beta_{1,a} + (\Sigma_{1,a} - \Sigma_{t,a})^{1/2}Z_{t}\\
            & \eps_{t} \in [0,1]
            \end{array}
\end{align}
The objective of~\eqref{eqn:opt} is differentiable in $\eps$, which motivates the use of stochastic gradient descent (SGD) to optimize the problem (see Algorithm~\ref{alg:gd}), where gradients can be efficiently computed via auto-differentiation libraries such as PyTorch~\cite{paszke2017automatic} or Jax~\cite{jax2018github}. Although gradient descent is not guaranteed to obtain the global maximizer, as \eqref{eqn:opt} is typically not concave, we find that in practice that SGD finds a good solution.

We also highlight that while we focus on cumulative regret as the core objective in this work, this framework is compatible with any other learning objective that is a differentiable function of the posterior parameters. The problem~\eqref{eqn:opt} can also incorporate constraints on exploration. A common constraint is setting a minimum level of exploration, e.g. $\eps_{t} \geq 0.1$, typically for the purpose of ensuring sufficient sample coverage for inference or off-policy evaluation~\cite{swaminathan2017off}.

Solving~\eqref{eqn:opt} under the prior $(\beta_{1}, \Sigma_{1})$ provides a principled way of obtaining the exploration schedule $\eps_{1:T}$. In addition, we can use the optimization problem as the basis for an~\emph{adaptive} exploration strategy. By re-solving the optimization problem~\eqref{eqn:opt} at every period $t$  (see Algorithm~\ref{alg:mpc}), the platform can adjust the exploration rate during the time horizon, increasing the exploration rate if the policy is learning slower than expected and decreasing it the policy has already converged. This control heuristic, known as model-predictive control (MPC), has been applied in domains such as chemical engineering, aeronautics, etc. but so far has not applied to the problem of tuning the exploration rate in contextual bandits.

\begin{algorithm}[t]
\caption{MPC}\label{alg:mpc}
\begin{algorithmic}
\Require Prior distribution $(\beta_{1}, \Sigma_{1})$, user embedding samples $\{X_{i} \}_{i=1}^{m}$, design matrix $I$, batch sizes $n_{t}$

\For{$t=1,...,T$}
\State Solve $\eps_{t:T}^{*} \gets \arg\min_{\eps_{t:T}} \overline{J}_{t}(\eps_{t:T}, \beta_{t},\Sigma_{t}).$
\Comment{Algorithm~\ref{alg:gd}}
\State Explore with rate $\eps_{t}^{*}$ in~\eqref{eqn:uniform_policy}.
\State Collect data $\mathcal{D}_{t} = \cup_{i=1}^{n_{t}}\{(X_{i},A_{i},R_{i},\xi_{i}):\xi_{i}=1\}$
\State Update posterior $(\beta_{t+1}, \Sigma_{t+1})$ using~\eqref{eqn:true_sigma_update}-\eqref{eqn:true_beta_update}
\EndFor
\end{algorithmic}
\end{algorithm}

\paragraph{Implementation Details}
While optimizing~\eqref{eqn:opt} is appealing due to its connection with regret minimization, doing so requires more inputs than simpler strategies. We proceed to discuss these inputs and how they can be set.
\begin{itemize}[leftmargin=0.5cm]
\item \textbf{Samples of user embeddings}. Samples of
$\{X_{i} \}_{i=1}^{m}$ are important for approximating expected rewards and for estimating the population design matrix $\mathcal{I}$. A simple way to obtain these are to first, set $\eps_{1} = 1$ (usually this is optimal) and then use the embeddings from users that arrive in the first batch $t = 1$. Alternatively, the platform could take a random sample of users and query their embeddings.
\item \textbf{Prior distribution.} While~\eqref{eqn:opt} seeks to minimize Bayesian regret under given prior, this prior may be misspecified. For improved performance, the platform can match the prior to the statistics of existing item embeddings. However, we find in experiments we find that even under a simple, misspecified prior ($\beta_{1} = \mathbf{0}, \Sigma_{1} = I$), the algorithms perform well.
\item \textbf{Batch sizes}. The optimization problem requires knowledge of the batch sizes $n_{t}$ across the horizon. Although it is unrealistic that the platform knows this exactly, these can be estimated from previous patterns of user traffic or impressions. In experiments, we find that our proposed algorithms are robust to errors in $n_{t}$, suggesting that $n_{t}$ does not need to be estimated perfectly to obtain benefits from optimization
\item \textbf{Diagonal Approximation}. The optimization problem involves repeated matrix inversions of the posterior covariance matrix, which can be computationally expensive. A more lightweight alternative is to replace $\Sigma_{t,a}$ and design matrix $\mathcal{I}$ with diagonal approximations for the posterior update, which we do for the experiments.
\end{itemize}  

\begin{figure*}[t]
\includegraphics[height=1.5in]{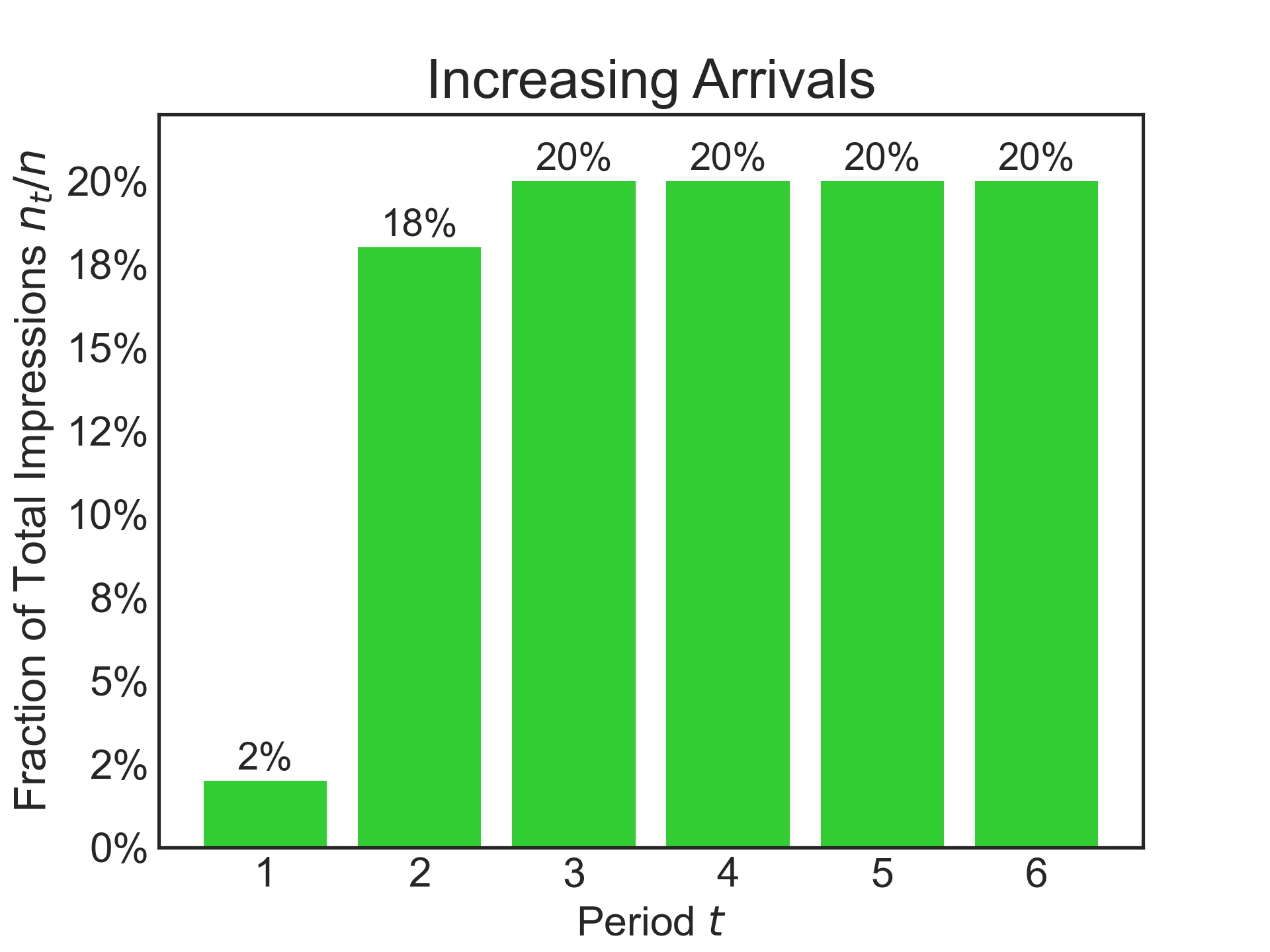}
\includegraphics[height=1.5in]{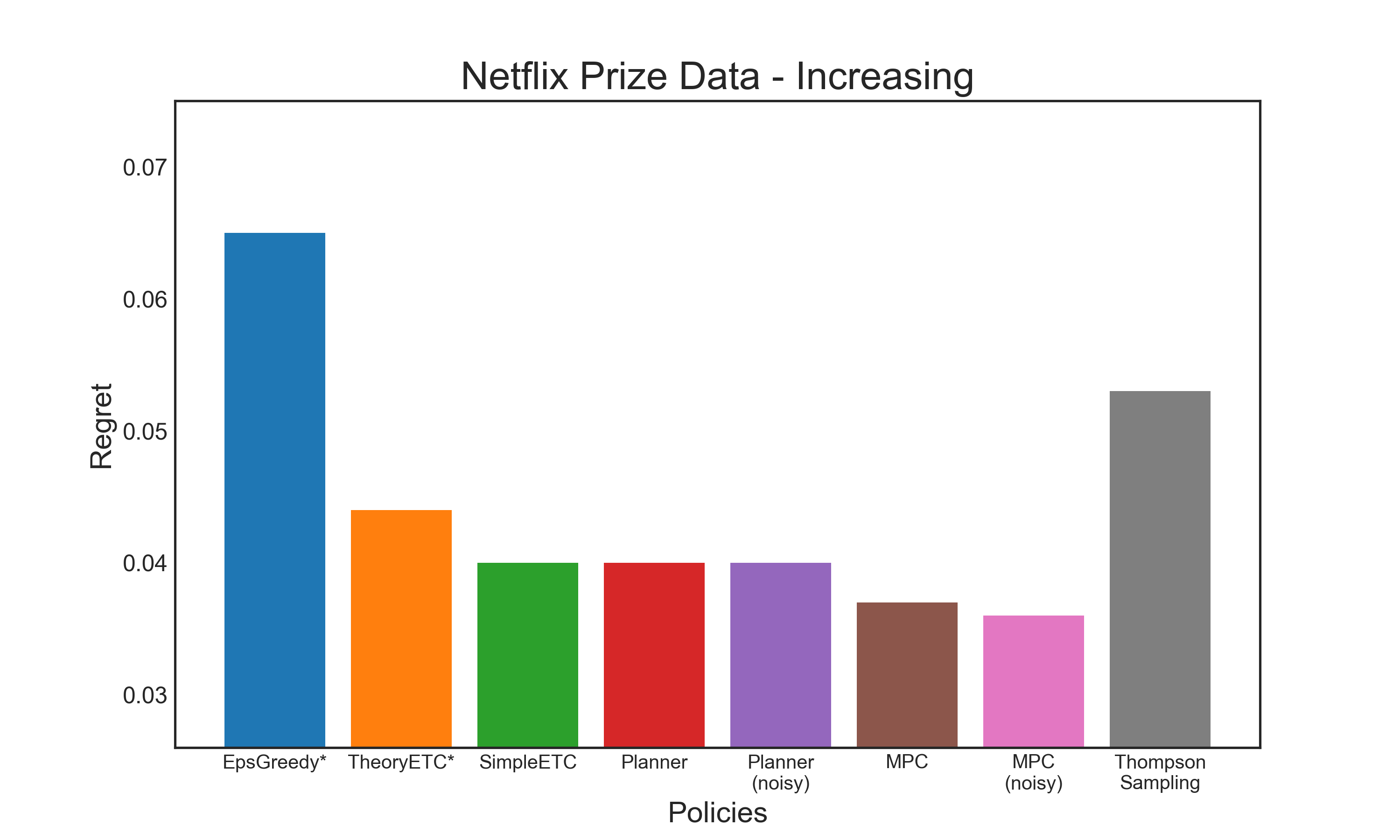}
\includegraphics[height=1.5in]{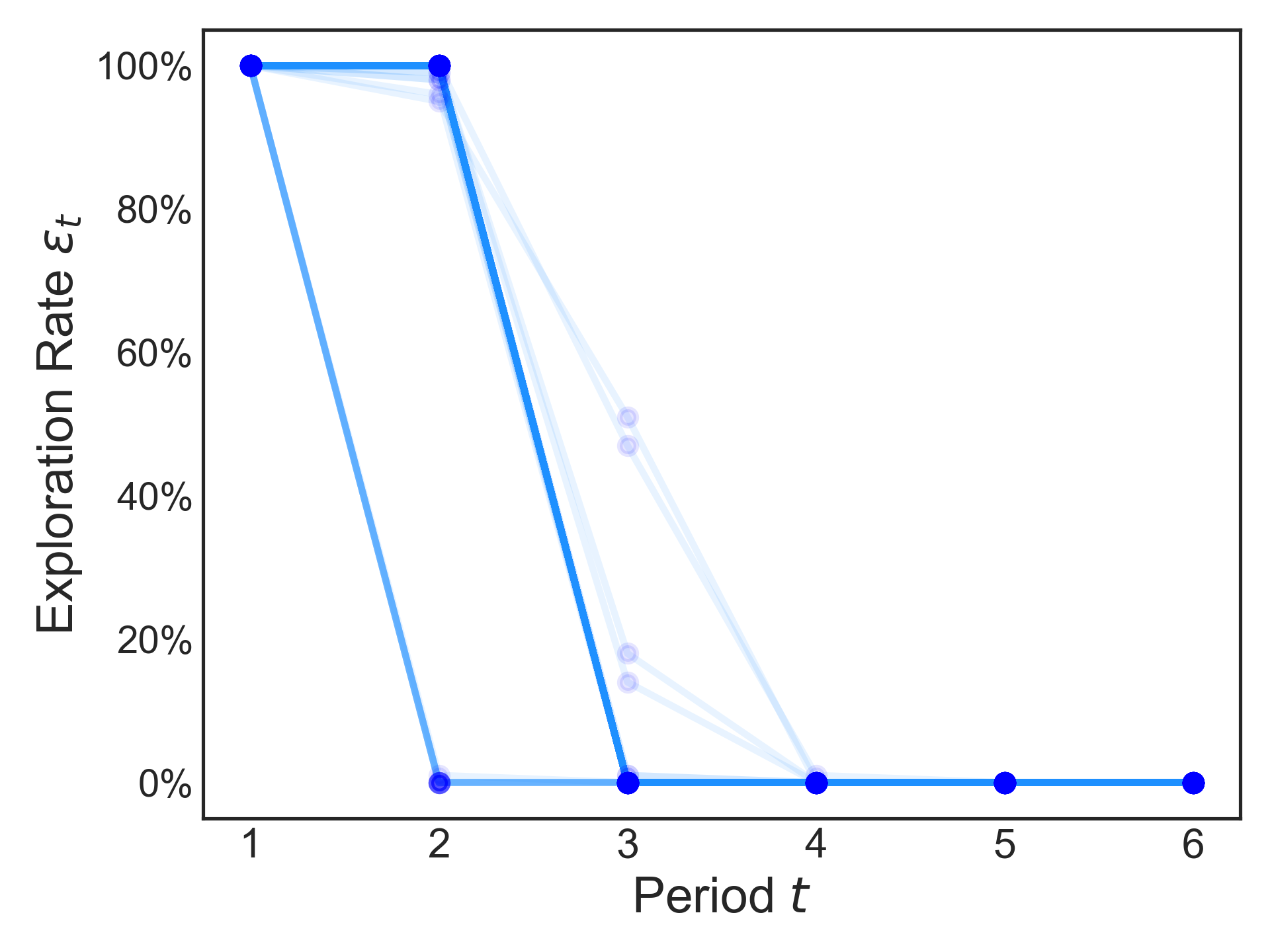}

\includegraphics[height=1.5in]{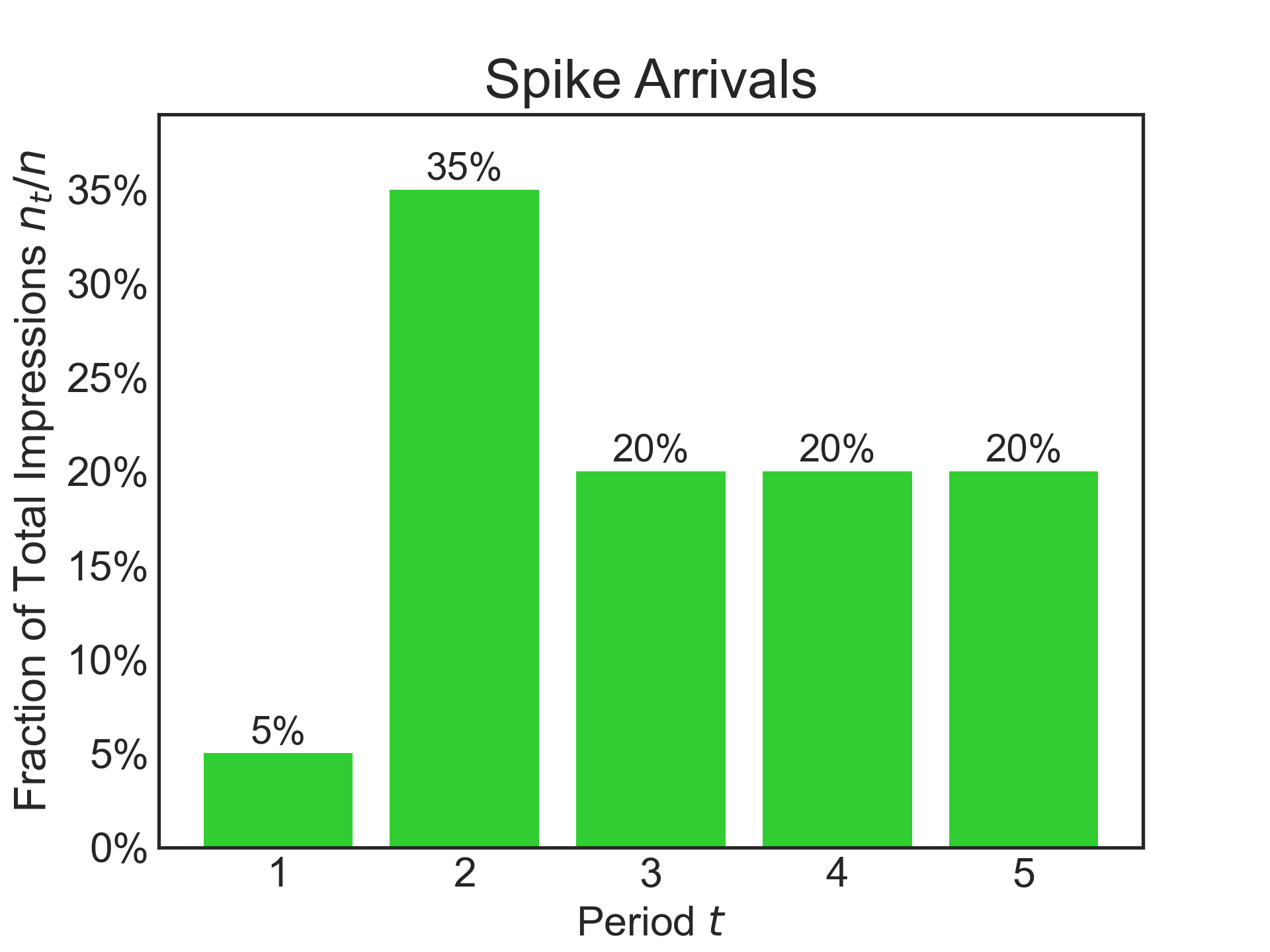}
\includegraphics[height=1.5in]{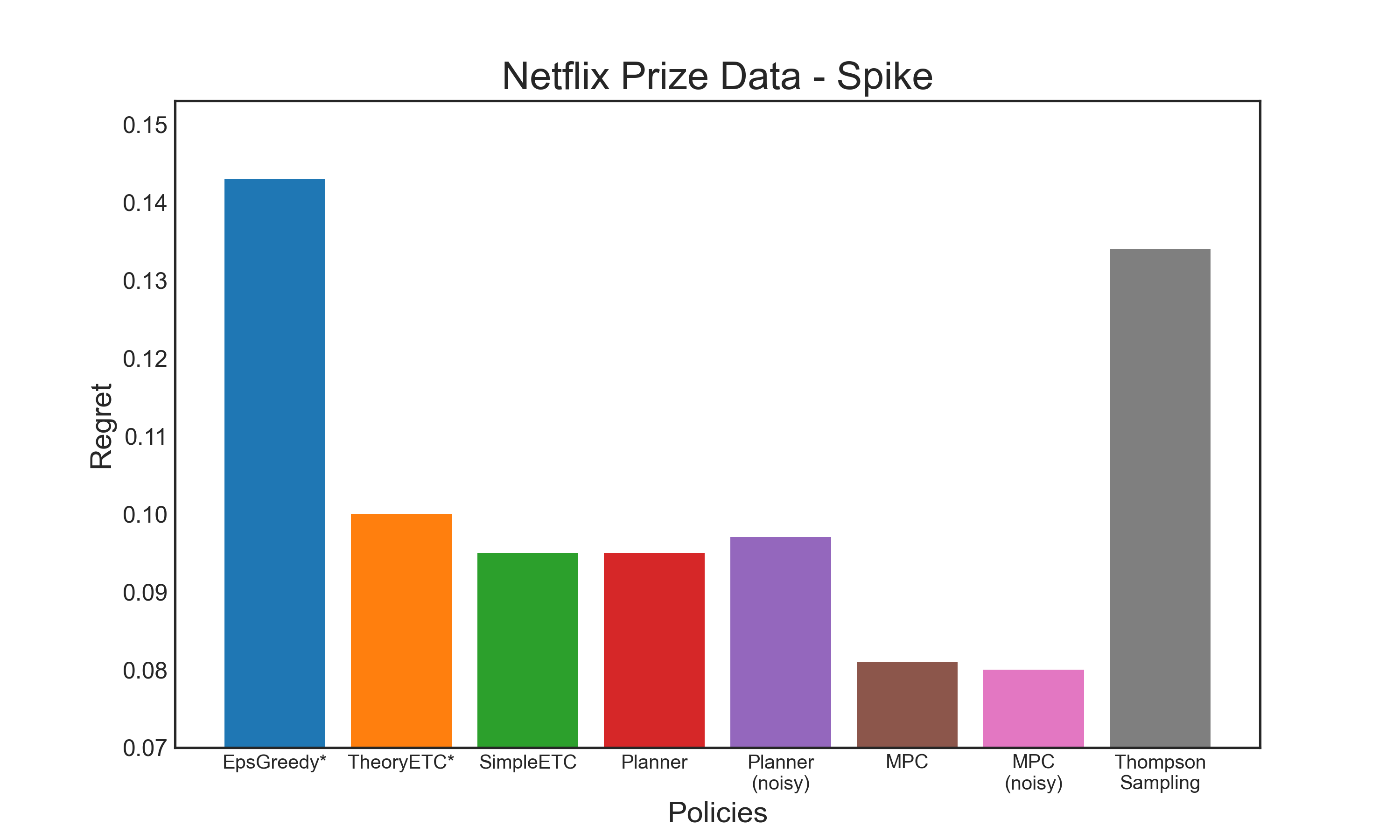}
\includegraphics[height=1.5in]{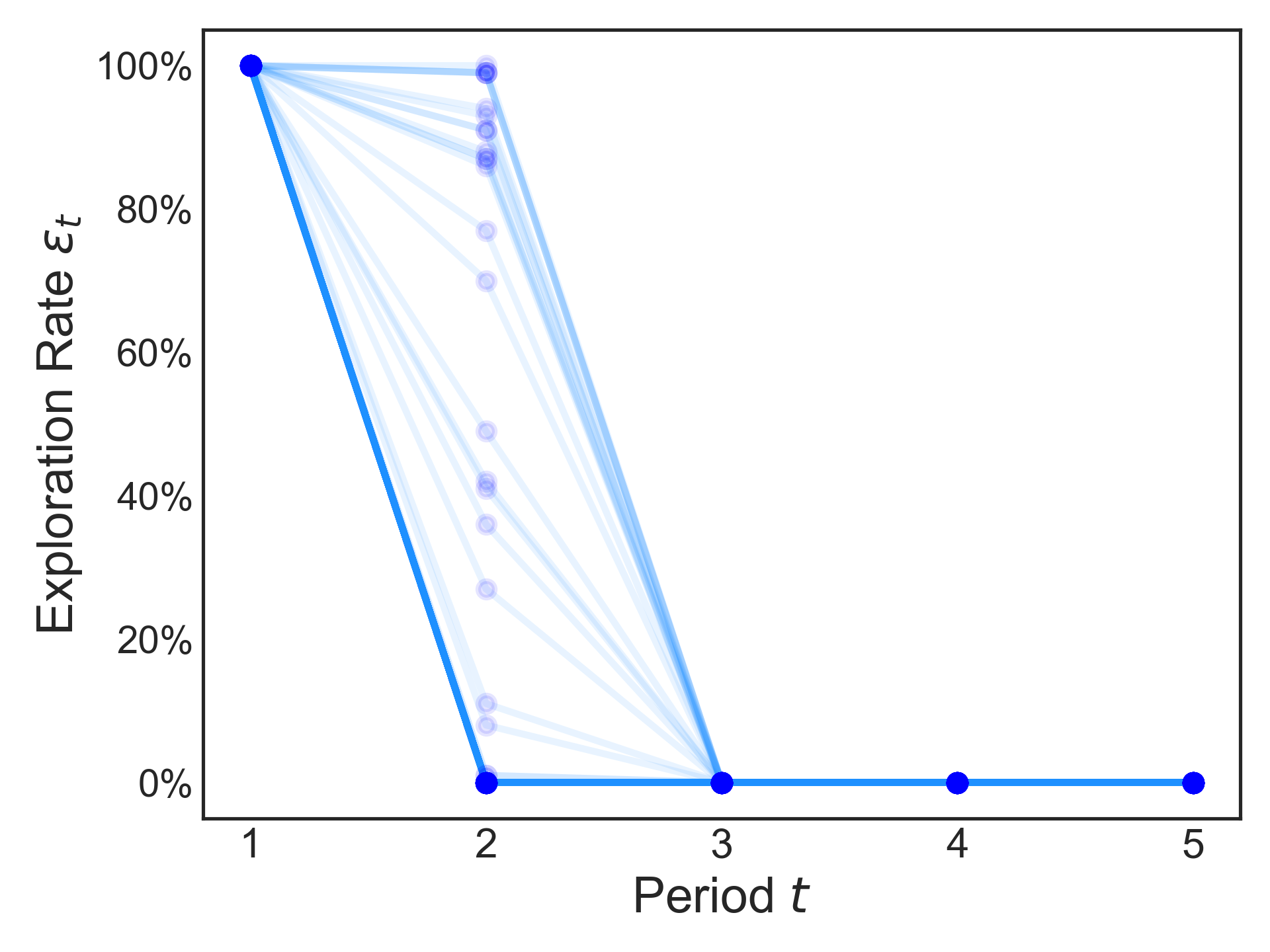}

\caption{\label{fig:exploration} (Left) Fraction of arrivals per period for the increasing and spike arrival patterns. (Middle) Average regret across methods for Netflix data with $K = 5$ arms and $N = 5000$ samples for increasing and spike arrival patterns. MPC (brown and pink) outperforms other methods in both scenarios (Right) Exploration rates $\eps_{t}$ from MPC for increasing (top) and spike (bottom) arrival patterns, where each line corresponds to one of 100 problem instances. Each line is the trajectory $\eps_{1:T}$ of exploration rates for a problem realization. Despite involving the same number of total samples, the exploration behavior is different for different arrival patterns, exploring mostly in period $t=1$ for the spike arrival pattern. We also see that MPC is adaptive, changing the exploration rates across different realizations, and often the final exploration rates differ from the initial plan.}
\end{figure*}

\section{Experiments}
\label{sec:experiments}

\begin{table*}[t]
\small

$K = 5$ items 
\vspace{0.1cm}

\begin{tabular}{lcccccc|cccccc}
\toprule
 & \multicolumn{6}{c|}{MovieLens-1M} & \multicolumn{6}{c}{Netflix} \\
\midrule
$N$ samples & \multicolumn{3}{c}{500} & \multicolumn{3}{c|}{5000} & \multicolumn{3}{c}{500} & \multicolumn{3}{c}{5000} \\
\midrule
Arrival $\lambda$ & Constant & Increasing & Spike & Constant & Increasing & Spike & Constant & Increasing & Spike & Constant & Increasing & Spike \\
\midrule
\textsf{EpsGreedy}$^{*}$ & 0.278 & 0.309 & 0.332 & 0.145 & 0.154 & 0.157 & 0.121 & 0.128 & 0.143 & 0.065 & 0.072 & 0.072 \\
\textsf{TheoryETC}$^{*}$ & 0.185 & \textbf{0.156} & \textbf{0.149} & 0.097 & 0.048 & 0.071 & \textbf{0.080} & 0.093 & 0.098 & 0.045 & 0.039 & 0.045 \\
\textsf{SimpleETC}       & 0.139 & 0.292 & 0.185 & \textbf{0.074} & 0.049 & \textbf{0.057} & 0.105 & 0.147 & 0.095 & \textbf{0.040} & 0.047 & \textbf{0.031} \\
\midrule
\textsf{Planner}         & 0.141 & 0.251 & 0.184 & \textbf{0.074} & 0.049 & \textbf{0.057} & 0.105 & 0.091 & 0.095 & \textbf{0.040} & 0.042 & \textbf{0.031} \\
\textsf{Planner (noisy)} & 0.136 & 0.185 & 0.187 & \textbf{0.074} & 0.051 & \textbf{0.057} & 0.107 & \textbf{0.083} & 0.097 & \textbf{0.040} & 0.044 & \textbf{0.031} \\
\textsf{MPC}             & \textbf{0.131} & 0.164 & 0.168 & \textbf{0.074} & 0.045 & \textbf{0.056} & 0.095 & \textbf{0.084} & \textbf{0.081} & \textbf{0.037} & \textbf{0.033} & \textbf{0.029} \\
\textsf{MPC (noisy)}     & \textbf{0.131} & \textbf{0.160} & 0.167 & \textbf{0.074} & \textbf{0.037} & \textbf{0.053} & 0.087 & 0.088 & \textbf{0.080} & \textbf{0.036} & \textbf{0.031} & \textbf{0.030} \\
\midrule
\textsf{TS}              & 0.253 & 0.287 & 0.351 & 0.178 & 0.189 & 0.258 & 0.101 & 0.113 & 0.134 & 0.053 & 0.061 & 0.097 \\
\bottomrule
\end{tabular}

\vspace{0.3cm}
$K = 10$ items 
\vspace{0.1cm}

\begin{tabular}{lcccccc|cccccc}
\toprule
& \multicolumn{6}{c|}{MovieLens-1M} & \multicolumn{6}{c}{Netflix} \\
\midrule
$N$ samples & \multicolumn{3}{c}{500} & \multicolumn{3}{c|}{5000} & \multicolumn{3}{c}{500} & \multicolumn{3}{c}{5000} \\
\midrule
Arrival $\lambda$ & Constant & Increasing & Spike & Constant & Increasing & Spike & Constant & Increasing & Spike & Constant & Increasing & Spike \\
\midrule
\textsf{EpsGreedy}$^{*}$ & 0.455 & 0.496 & 0.511 & 0.227 & 0.232 & 0.266 & 0.196 & 0.226 & 0.243 & 0.127 & 0.132 & 0.144 \\
\textsf{TheoryETC}$^{*}$ & 0.320 & 0.346 & \textbf{0.292} & 0.155 & 0.126 & 0.117 & 0.168 & 0.164 & 0.167 & 0.073 & \textbf{0.082} & 0.082 \\
\textsf{SimpleETC}       & 0.266 & 0.625 & 0.313 & \textbf{0.114} & 0.141 & 0.095 & 0.166 & 0.259 & 0.190 & \textbf{0.062} & 0.145 & \textbf{0.078} \\
\midrule
\textsf{Planner}         & 0.265 & 0.387 & 0.313 & \textbf{0.114} & 0.145 & 0.095 & 0.165 & 0.162 & 0.193 & \textbf{0.062} & 0.144 & \textbf{0.078} \\
\textsf{Planner (noisy)} & \textbf{0.255} & 0.368 & 0.324 & \textbf{0.114} & 0.137 & 0.095 & 0.164 & 0.172 & 0.181 & \textbf{0.062} & 0.130 & \textbf{0.078} \\
\textsf{MPC}             & 0.263 & \textbf{0.320} & \textbf{0.292} & \textbf{0.115} & \textbf{0.113} & 0.093 & 0.161 & 0.157 & \textbf{0.154} & \textbf{0.061} & 0.090 & \textbf{0.078} \\
\textsf{MPC (noisy)}     & 0.272 & 0.361 & \textbf{0.291} & \textbf{0.113} & 0.126 & \textbf{0.084} & \textbf{0.156} & \textbf{0.152} & 0.160 & \textbf{0.061} & \textbf{0.081} & \textbf{0.076} \\
\midrule
\textsf{TS}              & 0.462 & 0.547 & 0.651 & 0.312 & 0.327 & 0.404 & 0.204 & 0.251 & 0.247 & 0.111 & 0.124 & 0.197 \\
\bottomrule
\end{tabular}

\vspace{0.3cm}
$K = 15$ items 
\vspace{0.1cm}

\begin{tabular}{lcccccc|cccccc}
\toprule
& \multicolumn{6}{c|}{MovieLens-1M} & \multicolumn{6}{c}{Netflix} \\
\midrule
$N$ samples & \multicolumn{3}{c}{500} & \multicolumn{3}{c|}{5000} & \multicolumn{3}{c}{500} & \multicolumn{3}{c}{5000} \\
\midrule
Arrival $\lambda$ & Constant & Increasing & Spike & Constant & Increasing & Spike & Constant & Increasing & Spike & Constant & Increasing & Spike \\
\midrule
\textsf{EpsGreedy}$^{*}$ & 0.588 & 0.595 & 0.623 & 0.285 & 0.301 & 0.346 & 0.336 & 0.333 & 0.346 & 0.186 & 0.194 & 0.219 \\
\textsf{TheoryETC}$^{*}$ & 0.433 & 0.444 & \textbf{0.471} & 0.185 & 0.194 & 0.207 & \textbf{0.237} & \textbf{0.233} & 0.260 & 0.090 & \textbf{0.104} & 0.114 \\
\textsf{SimpleETC}       & \textbf{0.308} & 0.692 & 0.523 & \textbf{0.145} & 0.169 & \textbf{0.124} & 0.328 & 0.417 & 0.310 & \textbf{0.085} & 0.232 & \textbf{0.099} \\
\midrule
\textsf{Planner}         & \textbf{0.308} & \textbf{0.402} & 0.523 & \textbf{0.145} & 0.165 & \textbf{0.124} & 0.328 & 0.307 & 0.310 & \textbf{0.085} & 0.232 & \textbf{0.099} \\
\textsf{Planner (noisy)} & 0.315 & 0.425 & 0.501 & \textbf{0.145} & 0.167 & \textbf{0.124} & 0.327 & 0.302 & 0.306 & \textbf{0.085} & 0.230 & \textbf{0.099} \\
\textsf{MPC}             & \textbf{0.309} & 0.460 & \textbf{0.469} & \textbf{0.145} & \textbf{0.148} & \textbf{0.124} & 0.286 & 0.240 & \textbf{0.227} & \textbf{0.085} & 0.157 & \textbf{0.099} \\
\textsf{MPC (noisy)}     & \textbf{0.308} & 0.455 & 0.478 & \textbf{0.145} & 0.154 & \textbf{0.124} & 0.296 & \textbf{0.231} & \textbf{0.230} & \textbf{0.082} & 0.163 & \textbf{0.099} \\
\midrule
\textsf{TS}              & 0.525 & 0.615 & 0.660 & 0.278 & 0.306 & 0.360 & 0.290 & 0.328 & 0.404 & 0.209 & 0.219 & 0.259 \\
\bottomrule
\end{tabular}

\caption{\label{table:experiments} Average regret across contextual bandit settings. Each cell is averaged over 1000 bandit problems; standard errors are all close to $0.002$. While strategies like \textsf{TheoryETC}$^{*}$ or \textsf{SimpleETC} are good in some situations and poor in others, \textsf{Planner} and \textsf{MPC} match performance or even outperform across most scenarios, even under noisy estimates of batch sizes.}
\end{table*}

In this section, we investigate whether optimizing the objective in regret produces effective exploration schedules.

\paragraph{Bandit Environment}
To benchmark exploration strategies, we use two datasets to construct realistic contextual bandit problems that model cold-start exploration in recommendation systems
\begin{itemize}[leftmargin=0.5cm]
\item \textbf{Movie-Lens 1 Million data} (\emph{ML-1M}): consists of 6,000 users, 4,000 movies, and 1M interactions.
\item \textbf{Netflix Prize data} (\emph{Netflix}): we filter users and movies with less than 100 interactions, leaving with 28,604 users, 9,275 movies, and 3M interactions.
\end{itemize}

In a similar approach to~\cite{liang2022local}, we construct contextual bandit instances by first training a matrix factorization model~\cite{koren2009matrix} (embedding dimension $d=128$) on the data. We train the model to predict the rating, which we use as our reward metric.
We use the resulting user embeddings $x_{i}$ and item embeddings $\theta_{a}$ as the ground-truth reward model $r(x_{i},a)=x_{i}^\top \theta_{a}$. For each bandit instance, we randomly select a subset of $K \in \{5, 10, 15\}$ items from the total item set as the item set for the instance, denoted as $\mathcal{A}$. We run the contextual bandit policy for the time horizon of $T$ steps, starting from initial item embedding estimates of $\hat{\theta} = \mathbf{0} \in \R^{Kd}$ for the items in $\mathcal{A}$. At each period $t$, we sample a batch of $n_{t}$ users uniformly at random from the dataset. For a given user embedding $X_{i}$ and an item sampled by the policy $A_{i}$, we generate noisy rewards from the reward model $R_{i,a}=r(X_{i},A_{i}) + \eta_{i}$, where $\eta_{i} \sim N(0,1)$.

In order to assess the impact of batched feedback, we consider different patterns for the batch arrivals. We let $\lambda_{t}$ denote the fraction of users that arrive on period $t$, and we consider the following arrival patterns:
\begin{align*}
    \text{Increasing: } \lambda_{t} &= [2\%, 18\%, 20\%, 20\%, 20\%, 20\%] \\
    \text{Spike: }\lambda_{t} &= [5\%, 35\%, 20\%, 20\%, 20\%] \\
    \text{Constant: }\lambda_{t} &= 10\%\cdot \mathbf{1}_{10} 
\end{align*}
These patterns are motivated by user impressions in online platforms, which may experience predictable spikes (e.g. in the beginning of a new item launch) or may increase over time.
To generate random arrivals, we draw $n_{t} \sim \text{Bin}(N,\lambda_{t})$, where we consider $N \in \{500, 5000\}$. All the algorithms we benchmark do not know the realizations of $n_{t}$, but know $N$ and $\lambda_{t}$. For each choice of $K$, $N$, and arrival pattern $\lambda$, we run 1000 bandit problems, randomly sampling items and batch sizes $n_{t}$ each time, and report the regret~\eqref{eqn:regret} averaged across all problems.

\paragraph{Bandit Policies}

As discussed before, we focus on determining the exploration schedule for uniform exploration policies. Recall that these policies assign an item $a$ to a user $X_{i}$ according to,
\[
\label{eqn:uniform_policy}
\pi_{t}(a|X_{i}) =
\begin{cases}
1/K & \text{w.p. }  \eps_{t} \\
\mathbf{1}
\{
a = \arg \max_{a} X_{i}^{\top}\hat{\theta}_{t,a}
\}
& \text{w.p. } 1- \eps_{t}
\end{cases}
\]
where $\hat{\theta}_{t,a}$ is the Ridge regression estimator of $\theta_{a}$
\[
\hat{\theta}_{t,a} = \arg\min_{\theta \in \R^{d}} \|R_{t-1,a} - X_{t-1,a}\theta\|_{2}^{2} + \nu \|\theta\|_{2}^{2}
\]
where $\nu =1$ is the regularization factor,  $X_{t-1,a}$ is the matrix of all user-embeddings assigned $a$ before period $t$ and $R_{t-1,a}$ is the corresponding vector of observed rewards. Note that $\hat{\theta}_{t,a}$ is only estimated from users in the explore group, but we found that the results didn't qualitatively change if we included data from the exploit group as well. The evaluation metric is regret compared to the oracle optimal policy that assigns user $i$ the item $\arg \max_{a\in \mathcal{A}} r(X_{i},a)$.

We now discuss the exploration strategies we benchmark for choosing the exploration rates $\eps_{t}$.
\begin{itemize}[leftmargin=0.5cm]
\item \textbf{Epsilon-Greedy} [$\textsf{EpsGreedy}^{*}$]: constant exploration $\eps_{t} = \eps \in [0,1]$. Performance depends sensitively on $\eps$ so we report the regret of \emph{best} $\eps \in [0.05, 0.01, 0.1, 0.5, 1.0]$ for each problem instance (which is different across instances). This is a strong benchmark that is unrealistic to implement in practice, as it requires already knowing the optimal choice of $\eps$.
\item \textbf{Explore-then-Commit (ETC)} with budget $B = O(d^{2/3}N^{2/3})$ 
 [$\textsf{TheoryETC}^{*}$]. A theoretical recommendation for ETC is to use a budget of order $O(d^{1/3}N^{2/3})$~\cite{langford2007epoch, li2022simple, hao2020high}, and under mild assumptions, this results in a regret scaling of $O(N^{2/3})$. However, setting $B = d^{1/3}N^{2/3}$ tends to explore too much, so instead we report the regret for $B = cd^{1/3}N^{2/3}$ for the \emph{best} constant $c \in [0.05, 0.01, 0.1, 0.5, 1.0]$ (which is different across instances). This benchmark is also unrealistic to use in practice, as it requires already knowing the optimal $c$.
 \item \textbf{Simple ETC} [$\textsf{SimpleETC}$]: sets $\eps_{1} = 1$ and $\eps_{t} = 0$ for $t > 1$. Despite its simplicity, it is a commonly used heuristic.
 \end{itemize}

We propose the following exploration strategies based on optimization, and propose variants that test the robustness of these algorithms to errors in the input. 
\begin{itemize}[leftmargin=0.5cm]
 \item \textbf{Optimization Planner} [$\textsf{Planner}$]: Obtains the schedule via $\eps^{*}_{1:T} =\arg \min_{\eps_{1:T}} \overline{J}_{1}(\eps_{1:T},\beta_{1},\Sigma_{1})$ under (mis-specified) prior $\beta_{1} = \mathbf{0},\Sigma_{1} = I$. Obtains samples of $X_{i}$ from the first batch and uses the batch sizes $\bar{n}_{t} = N\lambda_{t}$ as inputs.
 \item \textbf{Optimization Planner (Noisy)} [$\textsf{Planner (noisy)}$]: Same as $\textsf{Planner}$ but uses noisy predictions of arrival rates $\widehat{n}_{t}=N\hat{\lambda}_{t}$ where $\hat{\lambda}_{t} \sim \text{Dirichlet}_{K\lambda}(\cdot)$ drawn from a Dirichlet distribution with parameter $(K\lambda_{1},...,K\lambda_{T})$.
 \item \textbf{Model-Predictive Control} [$\textsf{MPC}$]: Runs Algorithm~\ref{alg:mpc} with prior $\beta_{1} = \mathbf{0},\Sigma_{1} = I$, uses samples of $X_{i}$ from the first batch and uses batch sizes $\bar{n}_{t} = N\lambda_{t}$.
 \item \textbf{Model-Predictive Control (Noisy)} [$\textsf{MPC  (noisy)} $]: Same as \textsf{MPC}, but uses noisy predictions of arrival rates $\widehat{n}_{t}=N\hat{\lambda}_{t}$ where $\hat{\lambda}_{t} \sim \text{Dirichlet}_{K\lambda}(\cdot)$.
\end{itemize}
We also benchmark these uniform exploration strategies with a non-uniform exploration policy.
\begin{itemize}[leftmargin=0.5cm]
\item \textbf{Batched Thompson Sampling} [\textsf{TS}]~\cite{kalkanli2021batched, karbasi2021parallelizing}: At each period $t$, it samples item embeddings from the posterior  $\theta_{t} \sim N(\beta_{t},\Sigma_{t})$ and assigns items to each user in the batch according to $A_{i} = \arg \max_{a\in\mathcal{A}} X_{i}^{\top}\theta_{t,a}$. It then uses all the batch data to update the posterior mean and covariance.
\end{itemize}
\begin{table*}[t]
\small

$K = 10$ items 
\vspace{0.1cm}

\begin{tabular}{lcccccc|cccccc}
\toprule
 & \multicolumn{6}{c|}{MovieLens-1M} & \multicolumn{6}{c}{Netflix} \\
\midrule
$N$ samples & \multicolumn{3}{c}{500} & \multicolumn{3}{c|}{5000} & \multicolumn{3}{c}{500} & \multicolumn{3}{c}{5000} \\
\midrule
Arrival $\lambda$ & Constant & Increasing & Spike & Constant & Increasing & Spike & Constant & Increasing & Spike & Constant & Increasing & Spike \\
\midrule
\textsf{EpsGreedy}$^{*}$ & 0.456 & 0.494 & 0.511 & 0.227 & 0.232 & 0.267 & 0.197 & 0.225 & 0.243 & 0.127 & 0.133 & 0.144 \\
\textsf{TheoryETC}$^{*}$ & 0.349 & 0.350 & 0.373 & \textbf{0.158} & 0.167 & 0.170 & \textbf{0.159} & 0.169 & \textbf{0.171} & \textbf{0.075} & 0.092 & 0.087 \\
\textsf{SimpleETC}       & \textbf{0.278} & 0.436 & \textbf{0.306} & \textbf{0.157} & 0.130 & \textbf{0.130} & 0.168 & 0.228 & 0.192 & \textbf{0.075} & 0.102 & \textbf{0.082} \\
\midrule
\textsf{Planner}         & \textbf{0.280} & 0.411 & \textbf{0.305} & \textbf{0.157} & 0.130 & \textbf{0.130} & 0.168 & 0.170 & 0.191 & \textbf{0.075} & 0.102 & \textbf{0.082} \\
\textsf{Planner (noisy)} & 0.286 & 0.341 & \textbf{0.307} & \textbf{0.157} & 0.130 & \textbf{0.130} & 0.164 & 0.165 & 0.178 & \textbf{0.075} & 0.100 & \textbf{0.082} \\
\textsf{MPC}             & 0.285 & \textbf{0.326} & \textbf{0.308} & \textbf{0.157} & \textbf{0.126} & \textbf{0.130} & 0.169 & 0.163 & \textbf{0.171} & \textbf{0.074} & \textbf{0.081} & \textbf{0.081} \\
\textsf{MPC (noisy)}     & \textbf{0.282} & 0.350 & \textbf{0.305} & \textbf{0.157} & \textbf{0.125} & \textbf{0.130} & 0.165 & \textbf{0.158} & \textbf{0.167} & \textbf{0.074} & 0.087 & \textbf{0.081} \\
\bottomrule
\end{tabular}

\caption{\label{table:min_explore} Average regret for uniform exploration policies across contextual bandit settings with $K = 10$, where the exploration rate is constrained $\eps_{t}\geq0.05$ for all $t$. \textsf{Planner} and \textsf{MPC}, which incorporate the constraint into the optimization problem, reduce their exploration in the beginning as a response and more closely resemble \textsf{SimpleETC}.}
\end{table*}

\paragraph{Results}

Table~\ref{table:experiments} displays the average regret for each policy across different datasets and settings. We see that even when fixing the same total sample size $N$ for the horizon (in expectation), the distribution of arrivals $\lambda$ across periods has a large impact on regret and the optimal exploration strategy. Despite its simplicity, we find that \textsf{SimpleETC} often outperforms \textsf{TheoryETC}$^{*}$ for the Constant arrival pattern as well as the Spike arrival pattern (when $N = 5000$), despite the latter method's theoretical justification (and the grid search over optimal constants). For the Spike pattern, the batch size arriving on period $t = 2$ is large enough that it is better to be greedier at $t = 2$, even if this results in less data collected for future periods. However, \textsf{SimpleETC} is ineffective for the Increasing pattern because the first batch is too small to learn an effective policy and the strategy does not explore in future periods. This illustrates that under batched feedback, the optimal strategy can depend sensitively on the distribution of batch sizes, which makes it difficult to reason from first principles.

On the other hand, we observe that the optimization-based algorithms \textsf{Planner} and \text{MPC} are able to automatically tailor exploration to the structure of the batches. The exploration schedules consistently match and even outperform \textsf{SimpleETC} and \textsf{TheoryETC}$^{*}$ \emph{uniformly} across patterns, applying the strategy most appropriate to the situation. These algorithms do not overfit either; even under very noisy estimates of the true arrival pattern, both \textsf{Planner (noisy)} and \text{MPC (noisy)} achieve only slight worse performance. The consistent performance of this policy is promising for practitioners, as they can use the same algorithm for different settings, instead of having to know beforehand what the appropriate exploration strategy is. And when constraints are introduced, such as a minimum exploration rate of $\eps_{t}\geq 0.05$ in Table~\ref{table:min_explore}, incorporating these constraints into the optimization problem~\eqref{eqn:opt} induces less exploration in the beginning to account for this, delivering similar or better performance to the best heuristic in each setting. We also see that \textsf{MPC} often achieves lower regret than the static \textsf{Planner}, showing that adjusting the exploration rate during the horizon helps correct for any planning errors.

Finally, we note that surprisingly, the simple uniform exploration policies we consider consistently outperform Thompson Sampling in these batched, short-horizon settings. We find that Thompson Sampling suffers from over-exploration, which has also been observed in other works~\cite{jin2023thompson, do2024epsilon}. One driver of over-exploration is that Thompson Sampling lacks awareness of the time horizon, and explores until the posterior converges which may be sub-optimal for short time horizons. We do indeed find that time horizon is longer and when the arrival pattern is Constant, the performance gap closes. This illustrates though that seemingly simple and sub-optimal uniform exploration strategies can be surprisingly effective in settings relevant for practitioners.


\section{Conclusion}
\label{sec:conclusion}

In this work, we introduce a principled framework for selecting the exploration rates for uniform exploration policies in contextual bandit problems. This framework is based on optimization; selecting the exploration rates $\eps_{t}$ to minimize a differentiable approximation of Bayesian regret. We find that optimization is able to calibrate exploration automatically in a wide variety of scenarios and outperforms existing heuristics. This framework can be extended to include decisions other than exploration rate, such as potentially the allocation of traffic across periods $\lambda_{t}$ or allocation of items to users when exploring. It is also of interest to extend this to more complex reward models beyond linear models. The optimization problem could also include consideration for non-stationarity and additional constraints or objectives (e.g. Top-K regret). And while we focus on the bandit setting involving single item assignment, the optimization problem could augmented for slates of items~\cite{swaminathan2017off}.

\appendix
\section{Appendix}
\label{sec:appendix}

\paragraph{Proof of Proposition~\ref{thm:diff_regret}}

First, we can simplify the regret from random exploration
\begin{align*}
    \E_{1}[\text{Reg}(X_{i}, \mathbf{1}/K)] =&\E_{1}[r^{*}(X_{i})] - \frac{1}{K}\sum\nolimits_{a=1}^{K}\E_{1}[r(X_{i},a)] \\
     =& \E_{1}[r^{*}(X_{i})] - \frac{1}{K}\sum\nolimits_{a=1}^{K}\E[X_{i}^{\top}\E_{1}[\theta_{a}]] \\
     =&\E_{1}[r^{*}(X_{i})] - X_{i}^{\top}\bar{\beta}_{1}
\end{align*}
For the regret of the greedy policy, we have by the tower property of expectation, Bayesian regret of the greedy policy at period $t$ will be
\begin{align*}
    \E_{1}[\text{Reg}(X_{i}, \widehat{\pi}_{t})] =& \E_{1}[r^{*}(X_{i}) - \E_{A_{i} \sim \pi_{t}(\cdot |X_{i})} [r(X_{i},A_{i})]] \\
     =& \E_{1}[r^{*}(X_{i})] - \E_{1}[\E_{t}[\E_{A_{i} } [r(X_{i},A_{i})]]] \\
     =& \E_{1}[r^{*}(X_{i})] - \E_{1}[\max_{a} X_{i}^{\top}\beta_{t,a}]
\end{align*}
By Lemma~\ref{lemma:posterior}, we have that conditional on the prior, $\beta_{t,a}\stackrel{d}{=} \beta_{1,a} + (\Sigma_{1,a} - \Sigma_{t,a})^{1/2}Z_{t}$ where $Z_{t} \sim N(0,I)$ and $\Sigma_{t,a}$ is defined in~\eqref{eqn:sigma_update}.

\paragraph{Proof of Lemma~\ref{lemma:empirical}}

The first result follows from standard matrix Bernstein inequalities.
For simplicity of notation, assume that the measurement variance $s^{2}=1$.
\[
\sum_{l=t}^{s}I_{l,a}(\varepsilon_{l})=\sum_{l=t}^{s}\sum_{i=1}^{n_{t}}\xi_{t,i}\mathbf{1}\{A_{i}=a\}X_{t,i}X_{t,i}^{\top}\equiv\sum_{l=t}^{s}\sum_{i=1}^{n_{t}}M_{t,i}
\]
where $M_{t,i}=\xi_{t,i}\mathbf{1}\{A_{i}=a\}X_{t,i}X_{t,i}^{\top}$.
Since, $||X_{i}||_{2}^{2}\leq C$ we have that $||\xi_{t,i}\mathbf{1}\{A_{i}=a\}X_{t,i}X_{t,i}^{\top}\|_{\text{op}}\leq||X_{t,i}X_{t,i}^{\top}\|_{\text{op}}\leq C$.
Here we include time indices to distinguish between random variables
in different time periods. Note that this is a sum of iid random matrices
with mean equal to $I$ where $\varepsilon_{t}I=\varepsilon_{t}K^{-1}\mathbb{E}[X_{i}X_{i}^{\top}]$.
Using the matrix Bernstein inequality
\[
\mathbb{P}\left(\|\sum_{l=t}^{s}I_{l,a}(\varepsilon_{l})-\sum_{l=t}^{s}n_{l}\varepsilon_{l}I\|_{\text{op}}\geq t\right)\leq2d\exp\left(-\frac{t^{2}}{2(\sigma^{2}+t2C/3)}\right)
\]
where 
\begin{align*}
\sigma^{2} & =\|\sum_{l=t}^{s}\sum_{i=1}^{n_{t}}\mathbb{E}\left[(M_{t,i}-\varepsilon_{t}I)^{\top}(M_{t,i}-\varepsilon_{t}I)\right]\|_{\text{op}}\\
 & \leq\sum_{l=t}^{s}\sum_{i=1}^{n_{t}}\|\mathbb{E}\left[(M_{t,i}-\varepsilon_{t}I)^{\top}(M_{t,i}-\varepsilon_{t}I)\right]\|_{\text{op}}\\
 & \leq\sum_{l=t}^{s}\sum_{i=1}^{n_{t}}2\|\mathbb{E}\left[M_{t,i}^{\top}M_{t,i}\right]\|_{\text{op}}\\
 & \leq\sum_{l=t}^{s}\sum_{i=1}^{n_{t}}2C^{2}=n\lambda_{t:s}C^{2}
\end{align*}
where we use the shorthand $\lambda_{t:s}\equiv\sum_{l=t}^{s}\lambda_{s}$
Thus, for any $M>0$, we have that,
\begin{align*}
&\mathbb{P}\left(\|\sum_{l=t}^{s}I_{l,a}(\varepsilon_{l})-\sum_{l=t}^{s}n_{l}\varepsilon_{l}I\|_{\text{op}}\geq Mn^{1/2}\right) \\
&\leq2d\exp\left(-\frac{M^{2}n}{2(n\lambda_{t:s}C^{2}+n^{1/2}2C/3)}\right)
\end{align*}
Thus, for any $\delta>0$, we can find an $M$ such that the probability
is bounded by $\delta$.

\paragraph{Proof of Theorem~\ref{thm:approximation}}

While the theorem states the bound for $\bar{J}_{1}$, we prove a more general result for $\overline{J}_{t}$.
First, let $E_{n}:=\sum_{l=t}^{s}I_{n,l}(\varepsilon_{l})-\sum_{l=t}^{s}n_{l}\varepsilon_{l}I$.
For clarity, we let $S_{n}\equiv\sum_{l=t}^{s}I_{n,l}(\varepsilon_{l})$
and $S\equiv\sum_{l=t}^{s}\lambda_{l}\varepsilon_{l}I$ so that $\sum_{l=t}^{s}I_{n,l}(\varepsilon_{l})=nS+E_{n}$.
First, we have that, using the identity $A^{-1}-(A+B)^{-1}=(A+B)^{-1}BA^{-1}$
with $A=(\Sigma_{t,a}^{-1}+nS)$ and $B=E_{n}$, we have
\begin{align*}
 & \|\Sigma_{t,a}-(\Sigma_{t,a}^{-1}+nS)^{-1}-(\Sigma_{t,a}-(\Sigma_{t,a}^{-1}+nS+E_{n})^{-1})\|_{\text{op}}\\
 & =\|(\Sigma_{t,a}^{-1}+nS)^{-1}-(\Sigma_{t,a}^{-1}+nS+E_{n})^{-1}\|_{\text{op}}\\
 & =\|\left(\Sigma_{t,a}^{-1}+nS+E_{n}\right)^{-1}\left(E_{n}\right)\left(\Sigma_{t,a}^{-1}+nS\right)\|_{\text{op}}\\
 & \leq\|\left(\Sigma_{t,a}^{-1}+nS+E_{n}\right)^{-1}\|_{\text{op}}\|E_{n}\|_{\text{op}}\|\left(\Sigma_{t,a}^{-1}+nS\right)^{-1}\|_{\text{op}}
\end{align*}
We will let $K_{t}:=\|\left(\Sigma_{t,a}^{-1}+nS+E_{n}\right)^{-1}\|_{\text{op}}\|E_{n}\|_{\text{op}}\|\left(\Sigma_{t,a}^{-1}+nS\right)^{-1}\|_{\text{op}}$
in the last line. Note that
\begin{align*}
\|\left(\Sigma_{t,a}^{-1}+nS+E_{n}\right)^{-1}\|_{\text{op}} & =\lambda_{\max}\left(\left(\Sigma_{t,a}^{-1}+nS+E_{n}\right)^{-1}\right)\\
 & =\frac{1}{\lambda_{\min}\left(\left(\Sigma_{t,a}^{-1}+nS+E_{n}\right)\right)}\\
 & \leq\frac{1}{n\lambda_{\min}\left(S\right)+\lambda_{\min}(E_{n})}\\
 & \leq\frac{1}{n\lambda_{\min}\left(S\right)-||E_{n}||_{\text{op}}}
\end{align*}
and by a similar argument $\|\left(\Sigma_{t,a}^{-1}+nS\right)^{-1}\|_{\text{op}}\leq(n\lambda_{\min}(S))^{-1}$.
This gives us an upper bound of
\begin{align*}
 & \|(\Sigma_{t,a}^{-1}+nS)^{-1}-(\Sigma_{t,a}^{-1}+nS+E_{n})^{-1}\|_{\text{op}}\\
 & \leq\frac{||E_{n}||_{\text{op}}}{n\lambda_{\min}\left(S\right)-||E_{n}||_{\text{op}}}\frac{1}{n\lambda_{\min}(S)}
\end{align*}
Thus, we have that for any $M>0$, there exists $N$ such that for
all $n\geq N$,
\begin{align*}
 & \mathbb{P}\left(\|(\Sigma_{t,a}^{-1}+nS)^{-1}-(\Sigma_{t,a}^{-1}+nS+E_{n})^{-1}\|_{\text{op}}>Mn^{-3/2}\right)\\
 & \leq\mathbb{P}\left(\frac{||E_{n}||_{\text{op}}}{n\lambda_{\min}\left(S\right)-||E_{n}||_{\text{op}}}\frac{1}{n\lambda_{\min}(S)}>Mn^{-3/2}\right)\\
 & =\mathbb{P}\left(||E_{n}||_{\text{op}}>M\lambda_{\min}(S)n^{-1/2}(n\lambda_{\min}\left(S\right)-||E_{n}||_{\text{op}})\right)\\
 & =\mathbb{P}\left(||E_{n}||_{\text{op}}(1+M\lambda_{\min}(S)n^{-1/2})>M\lambda_{\min}(S)n^{-1/2}\right)\\
 & =O(e^{-M^{2}})
\end{align*}
Recall that $J_{t}$ is defined as,
\begin{align*}
J_{t} & =n\mathbb{E}_{t}[r^{*}(X_{i})]-\sum_{s=t}^{T}n_{s}\left(\varepsilon_{s}\mathbb{E}[X_{i}]^{\top}\bar{\beta}_{t}+(1-\varepsilon_{s})\mathbb{E}_{t}[\max_{a}X_{i}^{\top}\beta_{s,a}]\right)\\
 & \text{where }\beta_{s,a}=\beta_{t,a}+(\Sigma_{t,a}-\Sigma_{s,a})^{1/2}Z_{t,a}
\end{align*}
and $\overline{J}_{t}$ is identical with the replacement of $\beta_{s,a}$
by $\overline{\beta}_{s,a}$ where
\[
\overline{\beta}_{s,a}=\beta_{t,a}+(\Sigma_{t,a}-\overline{\Sigma}_{s,a})^{1/2}Z_{t,a}
\]
Note then that
\[
J_{t}-\overline{J}_{t}=\sum_{s=t}^{T}n_{s}(1-\varepsilon_{s})\left(\mathbb{E}_{t}[\max_{a}X_{i}^{\top}\beta_{s,a}]-\mathbb{E}_{t}[\max_{a}X_{i}^{\top}\overline{\beta}_{s,a}]\right)
\]
So one need only to bound the difference in $\mathbb{E}_{t}[\max_{a}X_{i}^{\top}\beta_{s,a}]-\mathbb{E}_{t}[\max_{a}X_{i}^{\top}\overline{\beta}_{s,a}]$.
First, we have that
\begin{align*}
X_{i}^{\top}\beta_{s,a} & \sim N(0,X_{i}^{\top}(\Sigma_{t,a}-\Sigma_{s,a})X)\\
X_{i}^{\top}\overline{\beta}_{s,a} & \sim N(0,X_{i}^{\top}(\Sigma_{t,a}-\overline{\Sigma}_{s,a})X)
\end{align*}
Thus, for $z_{a}\sim N(0,1)$ distributed iid we have
\begin{align*}
 & |\mathbb{E}_{t}[\max_{a}X_{i}^{\top}\beta_{s,a}]-\mathbb{E}_{t}[\max_{a}X_{i}^{\top}\overline{\beta}_{s,a}]|\\
 & =|\mathbb{E}_{X_{i}}\mathbb{E}_{z}[\max_{a}\sqrt{X_{i}^{\top}(\Sigma_{t,a}-\Sigma_{s,a})X_{i}}z_{a}-\max_{a}\sqrt{X_{i}^{\top}(\Sigma_{t,a}-\overline{\Sigma}_{s,a})X_{i}}z_{a}]|\\
 & \leq\mathbb{E}_{X_{i}}\mathbb{E}_{z}[|\max_{a}\sqrt{X_{i}^{\top}(\Sigma_{t,a}-\Sigma_{s,a})X_{i}}z_{a}-\max_{a}\sqrt{X_{i}^{\top}(\Sigma_{t,a}-\overline{\Sigma}_{s,a})X_{i}}z_{a}|]\\
 & =\mathbb{E}_{X_{i}}\mathbb{E}_{z}[\max_{a}|\sqrt{X_{i}^{\top}(\Sigma_{t,a}-\Sigma_{s,a})X_{i}}z_{a}-\sqrt{X_{i}^{\top}(\Sigma_{t,a}-\overline{\Sigma}_{s,a})X_{i}}z_{a}|]\\
 & \leq\mathbb{E}_{X_{i}}\mathbb{E}_{z}[\max_{a}|\sqrt{X_{i}^{\top}(\Sigma_{t,a}-\Sigma_{s,a})X_{i}}-\sqrt{X_{i}^{\top}(\Sigma_{t,a}-\overline{\Sigma}_{s,a})X_{i}}||z_{a}|]\\
 & \leq\mathbb{E}_{X_{i}}\mathbb{E}_{z}[\max_{a}|\sqrt{X_{i}^{\top}(\Sigma_{t,a}-\Sigma_{s,a})X_{i}-X_{i}^{\top}(\Sigma_{t,a}-\overline{\Sigma}_{s,a})X_{i}}||z_{a}|]\\
 & =\mathbb{E}_{X_{i}}\mathbb{E}_{z}[\max_{a}|\sqrt{X_{i}^{\top}(\overline{\Sigma}_{s,a}-\Sigma_{s,a})X_{i}}||z_{a}|]
\end{align*}
where we use the inequality $|\max_{a}x_{a}-\max_{a}y_{a}|\leq\max_{a}|x_{a}-y_{a}|$
and $|\sqrt{a}-\sqrt{b}|\leq\sqrt{|a-b|}$. Finally, we have that
\begin{align*}
 & \mathbb{E}_{X_{i}}\mathbb{E}_{t}[\max_{a}|\sqrt{X_{i}^{\top}(\overline{\Sigma}_{s,a}-\Sigma_{s,a})X_{i}}||z_{a}|]\\
 & \leq\mathbb{E}_{X_{i}}\max_{a}|\sqrt{X_{i}^{\top}(\overline{\Sigma}_{s,a}-\Sigma_{s,a})X_{i}}\cdot\mathbb{E}_{z}[\max_{a}|z_{a}|]\\
 & \leq\mathbb{E}_{X_{i}}\max_{a}|\sqrt{\|\overline{\Sigma}_{s,a}-\Sigma_{s,a}\|_{\text{op}}\|X_{i}\|_{2}^{2}}\cdot\mathbb{E}_{z}[\max_{a}|z_{a}|]\\
 & \leq\sqrt{C}\sqrt{\max_{a}\|\overline{\Sigma}_{s,a}-\Sigma_{s,a}\|_{\text{op}}}\cdot\sqrt{2\log2K}
\end{align*}
where we used a standard bound for $\mathbb{E}_{z}[\max_{a}|z_{a}|]\leq\sqrt{2\log2m}$.
This leaves us with
\[
|J_{t}-\overline{J}_{t}|\leq\sum_{s=t}^{T}n_{s}(1-\varepsilon_{s})\sqrt{2C\log2K}\sqrt{\max_{a}\|\overline{\Sigma}_{s,a}-\Sigma_{s,a}\|_{\text{op}}}
\]
By applying the probability bound in Lemma we have that for any $M>0$, 
\begin{align*}
\mathbb{P}\left(|J_{t}-\overline{J}_{t}|>Mn^{1/4}\right) & \leq\mathbb{P}\left(n\sqrt{2C\log2K}\sqrt{\max_{a}\|\overline{\Sigma}_{s,a}-\Sigma_{s,a}\|_{\text{op}}}>Mn^{1/4}\right)\\
 & =\mathbb{P}\left(\max_{a}\|\overline{\Sigma}_{s,a}-\Sigma_{s,a}\|_{\text{op}}>(2C\log2K)^{-1}Mn^{-3/2}\right)\\
 & \leq\sum_{a}\mathbb{P}\left(\|\overline{\Sigma}_{s,a}-\Sigma_{s,a}\|_{\text{op}}>(2C\log2K)^{-1}Mn^{-3/2}\right)\\
 & =O(Ke^{-M^{2}})
\end{align*}
By setting $M=O(\log(1/\delta))$, we have the desired result.

\bibliography{sizing_bib}
\bibliographystyle{abbrvnat}

\end{document}